\documentclass[smallextended]{svjour3} 

\usepackage{amsmath}
\usepackage{amssymb}
\usepackage{mathtools}
\usepackage{graphicx}
\usepackage[pdftex,dvipsnames]{xcolor}
\usepackage{xspace}
\usepackage{caption}
\usepackage{subcaption}
\usepackage{longtable}
\usepackage{booktabs}
\usepackage[lined,boxed,commentsnumbered, ruled,linesnumbered]{algorithm2e}
\usepackage{threeparttable}

\usepackage{natbib}  
\setcitestyle{authoryear,open={(},close={)}}

\usepackage{xargs} 
\usepackage[colorlinks=false]{hyperref} 

\newcommand{\ourmethod}{TS-CHIEF\xspace}
\newcommand{\ourmethodlong}{Time Series Combination of Heterogeneous and Integrated Embedding Forest}
\newcommand{\gitrepo}{https://github.com/dotnet54/TS-CHIEF}

\newcommand\ie[0]{\textit{i.e.} }

\newcommand{\reviewone}[1]{{\color{black}{}#1}}
\newcommand{\reviewtwo}[1]{{\color{black}{}#1}}

\DeclareMathOperator*{\argmax}{arg\,max}
\DeclareMathOperator*{\argmin}{arg\,min}

\def\BibTeX{{\rm B\kern-.05em{\sc i\kern-.025em b}\kern-.08emT\kern-.1667em\lower.7ex\hbox{E}\kern-.125emX}}

\usepackage{etoolbox}
\newcommand*{\affaddr}[1]{#1} 
\newcommand*{\affmark}[1][*]{\textsuperscript{#1}}

\begin{document}
\sloppy

\title{TS-CHIEF: A Scalable and Accurate Forest Algorithm for Time Series Classification}
\titlerunning{\ourmethod}        
\authorrunning{Shifaz et al.}

\author{Ahmed~Shifaz\affmark[1] \and Charlotte~Pelletier\affmark[1]\textsuperscript{,}\affmark[2] \and Fran\c{c}ois~Petitjean\affmark[1] \and Geoffrey~I.~Webb\affmark[1]}

 \institute{
     \affaddr{\affmark[1]Faculty of Information Technology}\\
       25 Exhibition Walk\\
       Monash University, Melbourne \\
       VIC 3800, Australia\\
      \affaddr{\affmark[2] IRISA, UMR CNRS 6074}\\
      Univ. Bretagne Sud\\
      Campus de Tohannic\\
      BP 573, 56 000 Vannes, France\\
      \email{
      \{ahmed.shifaz,francois.petitjean,geoff.webb\}@monash.edu,\\
      charlotte.pelletier@univ-ubs.fr}
 }
 
\maketitle


\begin{abstract}
Time Series Classification (TSC) has seen enormous progress over the last two decades. HIVE-COTE (Hierarchical Vote Collective of Transformation-based Ensembles) is the current state of the art in terms of classification accuracy. HIVE-COTE recognizes that time series data are a specific data type for which the traditional attribute-value representation, used predominantly in machine learning, fails to provide a relevant representation. HIVE-COTE combines multiple types of classifiers: each extracting information about a specific aspect of a time series, be it in the time domain, frequency domain or summarization of intervals within the series. However, HIVE-COTE (and its predecessor, FLAT-COTE) is often infeasible to run on even modest amounts of data. For instance, training HIVE-COTE on a dataset with only 1,500 time series can require 8 days of CPU time. It has polynomial runtime \reviewtwo{with respect to the} training set size, so this problem compounds as data quantity increases. \reviewone{We propose a novel TSC algorithm, \ourmethod (\ourmethodlong), which rivals HIVE-COTE in accuracy} but requires only a fraction of the runtime. \ourmethod constructs an ensemble classifier that integrates the most effective embeddings of time series that research has developed in the last decade. It~uses tree-structured classifiers to do so efficiently. We assess TS-CHIEF on 85 datasets of the University of California Riverside (UCR) archive, where it achieves state-of-the-art accuracy with scalability and efficiency. We demonstrate that TS-CHIEF can be trained on 130k time series in 2 days, a data quantity that is beyond the reach of any TSC algorithm with comparable accuracy. 


\end{abstract}




\section{Introduction}
\label{sec:introduction}

\reviewone{
    Time Series Classification (TSC) is an important area of machine learning research that has been growing rapidly in the past few decades \citep{Keogh2003, UCRArchive2018, Bagnall2017, Fawaz2018, yang200610, esling2012time,silva2018speeding}. 
    Numerous problems require classification of large quantities of time series data. These include land cover classification from temporal satellite images \citep{pelletier2019temporal}, human activity recognition \citep{nweke2018deep,wang2018deep}, classification of medical data from Electrocardiograms (ECG) \citep{wang2013bag},
    electric device identification from power consumption patterns \citep{Lines2015}, and many more \citep{rajkomar2018scalable,nwe2017convolutional, susto2018time}. The diversity of such applications are evident from the commonly used University of California Riverside (UCR) archive of TSC datasets \citep{Dau2018a, UCRArchive2015}.
    
    A number of recent TSC algorithms \citep{Lucas2018, Schafer2017a, Schafer2016} have tackled the issue of ever increasing data volumes, achieving greater efficiency and scalability than typical TSC algorithms. However, none has been competitive in accuracy to the state-of-the-art HIVE-COTE (Hierarchical Vote Collective of Transformation-based Ensembles) \citep{Lines2018}. 

    Our novel method, \ourmethod (\ourmethodlong), is a stochastic, tree-based ensemble that is specifically designed for speed and high accuracy. When building \ourmethod trees, at each 
    node we select from a random selection of TSC methods one that best classifies the data reaching the node. Some of these classification methods work with different representations of time series data \citep{Schafer2015, Bagnall2017}. Therefore, our technique combines decades of work in developing different classification methods for time series data \citep{Lucas2018, Lines2015, Schafer2015, Bagnall2016, Lines2018, Bagnall2017} and representations of time series data \citep{Bagnall2012, Bagnall2016, Schafer2015}, into a hetereogenous tree-based ensemble, that is able to capture a wide variety of discriminatory information from the dataset.
}

\reviewone{\ourmethod achieves scalability without sacrificing accuracy.}
It is orders of magnitude faster than HIVE-COTE \reviewone{(and its predecessor, FLAT-COTE)} while attaining \reviewone{a rank on accuracy} on the benchmark UCR archive \reviewone{that is almost indistinguishable}, as illustrated in Figure~\ref{fig:cd-us-vs-tsc} (on page \pageref{fig:cd-us-vs-tsc}). 

In addition, Figure~\ref{fig:scale-sat-time} (on page \pageref{fig:scale-sat-time}) shows an experiment that demonstrates the scalability of \ourmethod using the Satellite Image Time Series (SITS) dataset \citep{Tan2017}. It is 900x faster than HIVE-COTE for 1,500 time series (13~min \textit{versus} 8 days). 

Moreover, the relative speedup grows with data quantity: at 132k instances \ourmethod is 46,000x faster. For a training size that took \ourmethod 2 days, we estimated 234 years for HIVE-COTE.  

\reviewtwo{Overall, the following strategies are the key to attaining this exceptional efficiency without compromising accuracy: (1) using stochastic decisions during ensemble construction, (2) using stochastic selection instead of cross-validation for parameter selection, (3) using a tree-based approach to speed up training and testing, and (4) including improved variants of HIVE-COTE components Elastic Ensemble (EE) \citep{Bagnall2016}, Bag-of-SFA-Symbols (BOSS) \citep{Schafer2015} and Random Interval Spectral Ensemble (RISE) \citep{Lines2018}, but excluding its computationally expensive component Shapelet Transform (ST) \citep{Rakthanmanon2013b} (see Section~\ref{subsec:shapelets}).
}

The rest of the paper is organized as follows: Section~\ref{sec:relatedwork} discusses related work. Section~\ref{sec:method} presents our algorithm \ourmethod, \reviewone{and its time and space complexity}. In Section~\ref{sec:experiments}, we compare \reviewone{the accuracy of} \ourmethod against state-of-the-art TSC classifiers and investigate its scalability. \reviewone{In Section~\ref{sec:experiments}, we also study the variance of the ensemble, and the relative contributions of the ensemble's components.} Finally, in Section~\ref{sec:conclusion} we draw conclusions.

\section{Related Work}
\label{sec:relatedwork}
Time Series Classification (TSC) aims to predict a discrete label $y \in \{ 1, \cdots ,c \}$ for an unlabeled time series, where $c$ is the number of classes in the TSC task.
Although our work could be extended to time series with varying lengths and multi-variate time series, we focus here on univariate time series of fixed lengths.
A univariate time series $T$ of length $\ell$ is an ordered sequence of $\ell$ observations of a variable over time, where $T = \langle x_1, \cdots, x_\ell\rangle$, with $x_i \in \mathbb{R}$.
We use $D$ to represent a training time series dataset and $n$ to represent the number of time series in $D$.

\reviewtwo{We now present the main techniques used in TSC research. We also include a summary of training and test complexities of the methods present in this Section in Table~\ref{tab:complexities} (on page \pageref{tab:complexities}).}

\subsection{Similarity-based techniques}
\label{sec:wholeseries}

\sloppy These algorithms usually use 1-Nearest Neighbour (1-NN) with \textit{elastic} similarity measures. Elastic measures are designed to compensate for local distortions, miss-alignments or warpings in time series that might be due to stretched or shrunken
subsections within the time series. 

The classic benchmark for TSC has been 1-NN using Dynamic Time Warping (DTW), with cross validated warping window size \citep{Ding2008}. The warping window is a parameter that controls the \textit{elasticity} of the similarity measure. A~zero window size is equivalent to the Euclidean distance, while a larger warping window size allows points from one series to match points from the other series over longer time frames. 

Commonly used similarity measures include variations of DTW such as Derivative DTW (DDTW) \citep{Keogh2001a, Gorecki2013}, Weighted DTW (WDTW) \citep{Jeong2011}, Weighted DDTW (WDDTW) \citep{Jeong2011}, and measures based on edit distance such as Longest Common Subsequence (LCSS) \citep{Hirschberg1977}, Move-Split-Merge (MSM) \citep{Stefan2013}, Edit Distance with Real Penalty (ERP)\citep{Chen2004} and Time Warp Edit distance TWE \citep{Marteau2009}. Most of these measures have additional parameters that can be tuned. Details of these measures can be found in \citep{Lines2015, Bagnall2017}.

Ensembles formed using multiple 1-NN classifiers with a diversity of similarity measures have proved to be significantly more accurate than 1-NN with any single  measure \citep{Lines2015}. Such ensembles help to reduce the variance of the model and thus help to improve the overall classification accuracy. \reviewone{For example, Elastic Ensemble (EE) combines 11 1-NN algorithms, each using one of the 11 elastic measures \citep{Lines2015}.} For each measure, the parameters are optimized with respect to accuracy using cross-validation \citep{Lines2015, Bagnall2017}. Though EE is a relatively accurate classifier \citep{Bagnall2017}, it is slow to train due to high computational cost of the leave-one-out cross-validation used to tune its parameters -- \reviewone{$O(n^2 \cdot \ell^2)$}. Furthermore, since EE is an ensemble of 1-NN models, the classification time for each time series is also high -- $O(n \cdot \ell ^2)$.

Our recent contribution, Proximity Forest (PF), is more scalable and accurate than EE \citep{Lucas2018}. It builds an ensemble of classification trees, where data at each node are split based on  similarity to a representative time series from each class.  This contrasts with the standard attribute-value splitting methods used in decision trees. 
Degree of similarity is computed by selecting at random one measure among the 11 used in EE. The parameters of the measures are also selected at random.
Proximity Forest is highly scalable owing to the use of a divide and conquer strategy, and stochastic   parameter selection in place of computationally expensive parameter tuning. 

\subsection{Interval-based techniques}

These algorithms select a set of intervals from the whole series and apply transformations to these intervals to generate a new feature vector. The new feature vector is then used to train a traditional machine learning algorithm, \reviewone{usually a forest of Random Trees, similar to Random Trees used in Random Forest (but without bagging)}. For instance, Time Series Forest (TSF) \citep{Deng2013} applies three time domain transformations --~mean, standard deviation and slope~-- to each of a set of randomly chosen intervals, and then trains a decision tree using this new data representation. The operation is repeated to learn an ensemble of decision trees, similar to \reviewone{Random Trees}, on different randomly chosen intervals. Other notable interval-based algorithms are Time Series Bag of Features (TSBF) \citep{Baydogan2013}, Learned Pattern Similarity (LPS) \citep{Baydogan2016}, and the recently introduced Random Interval Spectral Ensemble (RISE) \citep{Lines2018}. 

RISE computes four different transformations for each random interval selected: Autocorrelation Function (ACF), Partial Autocorrelation Function (PACF), and Autoregressive model (AR) which extracts features in time domain, and Power Spectrum (PS) which extracts features in the frequency domain \citep{Lines2018, Bagnall2016}. Coefficients of these functions are used to form a new transformed feature vector. \reviewone{After these transformations have been computed for each interval, a Random Tree is trained on each of the transformed intervals. The training complexity of RISE is $O(k \cdot n \cdot \ell^2)$ \citep{Lines2018}, and the test complexity is $O(k \cdot log(n) \cdot \ell^2)$.}

The algorithm presented in this paper has components inspired by RISE, therefore, further details are presented later (see Section ~\ref{subsec:interval}). 

\subsection{Shapelet-based techniques}
\label{subsec:shapelets}

Rather than extracting intervals, where the location of sub-sequences are important, shapelet-based algorithms seek to identify sub-sequences that allow discrimination between classes irrespective of where they occur in a sequence \citep{Ye2009}. Ideally, a good shapelet candidate should be a sub-sequence similar to time series from the same class, and dissimilar to time series from other classes. Similarity is usually computed using the minimum Euclidean distance of a shapelet to all sub-sequences of the same length from another series. 

The original version of the shapelet algorithm \citep{Ye2009, Mueen2011}, enumerates all possible sub-sequences among the training set to find the ``best'' possible shapelets. It uses Information Gain criteria to asses how well a given shapelet candidate can split the data. The ``best'' shapelet candidate and a distance threshold is used as a decision criterion at the node of a binary decision tree. The search for the ``best'' shapelet is then recursively repeated until obtaining pure leaves. Despite some optimizations proposed in the paper, it is still a very slow algorithm with training complexity of $O(n^2 \cdot \ell^4)$. 

Much of the research about shapelets has focused on ways of speeding up the shapelet discovery phase. Instead of enumerating all possible shapelet candidates, researchers have tried to come up with ways of quickly identifying possible ``good'' shapelets. These include Fast Shapelets (FS) \citep{Rakthanmanon2013a} and Learned Shapelets (LS) \citep{Grabocka2014}. Fast Shapelet proposed to use an approximation technique called Symbolic Aggregate Approximation (SAX) \citep{Lin2007} to shorten the time series during the shapelet discovery process in order to speed up by reducing the number of shapelet candidates. Learned Shapelets (LS) attempted to ``learn'' the shapelets rather than enumerate all possible candidates. Fast Shapelets algorithm is faster than LS, but it is less accurate \citep{Bagnall2017}.

Another notable shapelet algorithm is Shapelet Transform (ST) \citep{Hills2014}. In~ST, the `best' $k$ shapelets are first extracted based on their ability to separate classes using a quality measure such as Information Gain, and then the distance of each of the ``best'' $k$ shapelets to each of the samples in the training set is computed \citep{Hills2014, Bostrom2015, Large2017}.
The distance from $k$ shapelets to each time series forms a matrix of distances which defines a new transformation of the dataset. \reviewone{This transformed dataset is finally used to train an ensemble of eight traditional classification algorithms including 1-Nearest Neighbour with Euclidean distance and DTW, C45 Decision Trees, BayesNet, NaiveBayes, SVM, Rotation Forest and Random Forest. Although very accurate, ST also has a high training-time complexity of $O(n^2 \cdot \ell^4)$ \citep{Hills2014, Lines2018}.}

One algorithm that speeds up the shapelet-based techniques is Generalized Random Shapelet Forest (GRSF) \citep{Karlsson2016}. GRSF selects a set of random shapelets at each node of a decision tree and performs the shapelet transformation at the node level of the decision tree. GRSF is fast because it is tree-based and uses random selection of shapelets instead of enumerating all shapelets. GRSF experiments were carried out on a subset of the 85 UCR datasets where the values of the hyperparameters --~the number of randomly selected shapelets as well as the lower and upper shapelet lengths~-- are optimized by using a grid search.

\subsection{Dictionary-based techniques}

Dictionary-based algorithms transform time series data into bag of words \citep{Senin2013, Schafer2015, Large2018}. Dictionary based algorithms are good at handling noisy data and finding discriminatory information in data with recurring patterns \citep{Schafer2015}. Usually, an approximation method is first applied to reduce the length of the series \citep{Keogh2001b, Lin2007, Schafer2012}, and then a quantization method is used to discretize the values, and thus to form words \citep{Schafer2015, Large2018}. Each time series is then represented by a histogram that counts the word frequencies. 1-NN with a similarity measure, that compares the similarity between histograms, can then be used to train a classification model. Notable dictionary based algorithms are Bag of Patterns (BoP) \citep{Lin2012}, Symbolic Aggregate Approximation-Vector Space Model (SAX-VSM) \citep{Senin2013}, Bag-of-SFA-Symbols (BOSS) \citep{Schafer2015}, BOSS in Vector Space (BOSS-VS) \citep{Schafer2016} and Word eXtrAction for time SEries cLassification (WEASEL) \citep{Schafer2017a}.

To compute an approximation of a series, BOP and SAX-VSM use a method called Symbolic Aggregate Approximation (SAX) \citep{Lin2007}. SAX uses Piecewise Aggregate Approximation (PAA) \citep{Keogh2001b} which concatenates the means of consecutive segments of the series and uses quantiles of the normal distribution as breakpoints to discretize or quantize the series to form a word representation. By contrast, BOSS, BOSS-VS, and WEASEL use a method called Symbolic Fourier Approximation (SFA) \citep{Schafer2012} to compute the approximated series. SFA applies Discrete Fourier Transformation (DFT) on the series and uses the coefficients of DFT to form a short approximation, representing the frequencies in the series. This approximation is then discretized using a data-adaptive quantization method called Multiple Coefficient Binning (MCB) \citep{Schafer2012, Schafer2015}.

\reviewone{The most commonly used algorithm in this category is Bag-of-SFA-Symbols (BOSS), which is an ensemble of dictionary-based 1-NN models \citep{Schafer2015}. BOSS is a component of HIVE-COTE and our algorithm also has a component inspired by BOSS.}
Further details of the BOSS algorithm will be presented in Section~\ref{sec:method}. BOSS has a training time complexity of $O(n^2 \cdot \ell^2)$ \reviewone{and a testing time complexity of $O(n \cdot \ell)$} \citep{Schafer2015}. \reviewone{A~variant of BOSS called BOSS-VS \citep{Schafer2016} has a much faster train and test time while being less accurate. The more recent variant WEASEL \citep{Schafer2017a} is more accurate but has a slower training time than BOSS and BOSS-VS, in addition to high space complexity \citep{Schafer2017a,Lucas2018, middlehurst2019scalable}.}

\subsection{Combinations of Transformations}

Two leading algorithms that combine multiple transformations are Flat Collective of Transformation-Based Ensembles (FLAT-COTE) \citep{Bagnall2016} and the more recent variant Hierarchical Vote COTE (HIVE-COTE) \citep{Lines2018}. FLAT-COTE is a meta-ensemble of 35 different classifiers that use different time series classification methods such as similarity-based, shapelet-based, and interval-based techniques. In particular, it includes other ensembles such as EE and ST. The label of a time series is determined by applying weighted majority voting, where the weighting of each constituent depends on the training leave-one-out cross-validation (LOO CV) accuracy. HIVE-COTE works similarly, but it includes new algorithms, BOSS and RISE, and changes the weighted majority voting to make it balance between each type of constituent module. These modifications result in a major gain in accuracy, and it is currently considered as the state of the art in TSC for accuracy. However, both variants of COTE have high training complexity, lower bounded by the slow cross-validation used by EE -- \reviewone{$O(n^2 \cdot \ell^2)$} 
-- and exhaustive shapelet enumeration used by ST -- $O(n^2\cdot \ell^4)$.

\subsection{Deep Learning}

Deep learning is interesting for time series both because of the structuring dimension offered by time (deep learning has been particularly good for images and videos) and for its linear scalability with training size. Most related research has focused on developing specific architectures based mainly on Convolutional Neural Networks (CNNs) \citep{Wang2017, Fawaz2018},
coupled with data augmentation, which is required to make it possible for them to reach high accuracy on the relatively small training set sizes present in the UCR archive \citep{le2016data, Fawaz2018}. While these approaches are computationally efficient, the two leading algorithms, Fully Connected Network (FCN) \citep{Wang2017} and Residual Neural Network (ResNet) \citep{Wang2017}, are still less accurate than FLAT-COTE and HIVE-COTE \citep{Fawaz2018}.

\section{\ourmethod}
\label{sec:method}

This section introduces our novel algorithm \ourmethod, which stands for \textbf{T}ime \textbf{S}eries  \textbf{C}ombination of \textbf{H}eterogeneous and \textbf{I}ntegrated \textbf{E}mbeddings \textbf{F}orest.
\ourmethod is an ensemble algorithm that makes the most of the scalability of tree classifiers coupled with the accuracy  brought by decades of research into specialized techniques for time series classification. Traditional attribute-value decision trees form a tree by recursively splitting the data \reviewtwo{with respect to} the value of a selected attribute. These techniques (and ensembles thereof) do not in general perform well when applied directly to time series data \citep{Bagnall2017}. As they treat the value at each time step as a distinct attribute, they are unable to exploit the information in the series order. In contrast, \ourmethod utilizes splitting criteria that are specifically developed for time series classification.

Our starting point for \ourmethod is the Proximity Forest (PF) algorithm \citep{Lucas2018}, which builds an ensemble of classification trees with `splits' using the proximity of a given time series $T$ to a set of reference time series: if $T$ is closer to the first reference time series, then it goes to the first branch, if it is closer to the second reference time series, then it goes to the second branch, and so on. Proximity Forest integrates 11 time series measures for evaluating similarity.
At each node a set of reference series is selected, one per class, together with a similarity measure and its parameterization. These selections are made stochastically.
Proximity Forest attains accuracies that are comparable to BOSS and ST (see Figure~\ref{fig:cd-us-vs-tsc}).
\ourmethod  complements Proximity Forest's splitters with dictionary-based and interval-based splitters, which we describe below. Our algorithmic contributions are three-fold: 
\begin{enumerate}
    \item We take the ideas that underlie the best dictionary-based method, BOSS, and develop a tree splitter based thereon.
    \item We take the ideas behind the best interval-based method, RISE, and develop a tree splitter based thereon.
    \item We develop techniques to integrate these two novel splitters together with those introduced by Proximity Forest, such that any of the 3 types might be used at any node of the tree.
\end{enumerate}
\ourmethod is an ensemble method: we thus paid particular attention to maximizing the diversity between the learners in its design. 
We~do this by creating a very large space of possible splitting criteria. This diversity for diversity sake would be unreasonable if the objective was to create a single standalone classifier. By contrast, by ensembling, this diversity can be expected to reduce the covariance term of ensemble theory \citep{Ueda1996}. If ensemble member classifiers are too similar to one another, their collective decision will differ little from that of a single member.

\subsection{General Principles}

During the training phase, \ourmethod builds a forest of $k$ trees. 
The general principles of decision trees remain: tree construction starts from the root node and recursively builds the sub-trees, and at each node, the data is split into branches using a splitting function. 
Where \ourmethod differs is in the use of time-series-specific splitting functions. 
The details of these splitting functions will be discussed in Section~\ref{sec:splitfunctions}. In short, we use different types of splitters either using time series similarity measures, dictionary-based or interval-based representations. At each node, we generate a set of \emph{candidate} splits and select the best one using the weighted Gini index, \ie{}the split that maximizes the purity of the created branches (similar to a classic decision tree). 
We describe the top-level algorithm in Algorithm~\ref{alg:build-tree}; note that this algorithm is very typical of decision trees and that all the time-series-specific features are in the way we generate candidate splits, as shown in Algorithms~\ref{alg:gen-similarity-split}, \ref{alg:gen-dictionary-split}
and \ref{alg:select-interval-split}.

\begin{algorithm}
\caption{$\mathrm{build\_tree}(D,C_e,C_b,C_r$)}
\label{alg:build-tree}
\DontPrintSemicolon
\KwIn{$D$: a time series dataset}
\KwIn{$C_e$: no.~of similarity-based candidates}
\KwIn{$C_b$: no.~of dictionary-based candidates}
\KwIn{$C_r$: no.~of interval-based candidates}
\KwOut{$T$: a \ourmethod Tree}
\vspace*{5pt}

\uIf{$\mathrm{is\_pure}(D)$}{\Return $\mathrm{create\_leaf}(D)$\;}

$T \leftarrow \mathrm{create\_node}()$  \hspace{0.5cm} \tcp*{Create tree represented by its root node}

$\mathcal{S} \leftarrow \emptyset$ \hspace{1cm}\tcp*{set of candidate splitters}

\vspace*{5pt}
$\mathcal{S}_e \leftarrow \mathrm{generate\_similarity\_splitters}(D, C_e)$\;
Add all similarity-based splitters in $\mathcal{S}_e$ to $\mathcal{S}$\;

\vspace*{5pt}
$\mathcal{S}_b \leftarrow \mathrm{generate\_dictionary\_splitters}(D, C_b)$\;
Add all dictionary-based splitters in $\mathcal{S}_b$ to $\mathcal{S}$\;

\vspace*{5pt}
$\mathcal{S}_r \leftarrow \mathrm{generate\_interval\_splitters}(D, C_r)$\;
Add all interval-based splitters in $\mathcal{S}_r$ to $\mathcal{S}$\;

\vspace*{5pt}
$\delta^\star\leftarrow \argmax\limits_{\delta\in \mathcal{S}}  \mathrm{Gini}\left(\delta\right)$ \tcp*{select the best splitter using Gini}\;

$T_\delta \leftarrow \delta^\star$ \hspace{1cm} \tcp*{store the best splitter in the new node $T$}
$T_B\leftarrow \emptyset$ \hspace{1cm} \tcp*{store the set of branch nodes in $T$}

\tcp{Partition the data using $\delta^\star$ and recurse}

\uIf{$\delta^\star$ \emph{is similarity-based}}{
  \ForEach{$e \in \delta^\star_E$}{
  
    \tcp{$\delta^\star_M$ is the distance measure of the best similarity-based splitter $\delta^\star$ selected by Gini}
  
    $D^{+}\leftarrow\{d\in D\mid \delta^\star_M(d,e)=\min_{x\in \delta^\star_E}(\delta^\star_M(d,x))$\; 

    $t_e\leftarrow\mathrm{build\_tree}(D^{+},C_e,C_b,C_r)$\;
    
    Add new branch $t_e$ to $T_B$\;
  }
}
\uElseIf{$\delta^\star$ \emph{is dictionary-based}}{
  \ForEach{$e \in \delta^\star_E$}{
  
    \tcp{For definition of \text{BOSS\_dist}, see \cite[Definition 4]{Schafer2015}} 
    
    \tcp{$\delta^\star_{\mathcal{T}}(d)$ is the BOSS transformation of $d$ using the BOSS transform function $\delta^\star_{\mathcal{T}}$ of the best dictionary-based splitter $\delta^\star$ selected by Gini}
    
    $D^{+}\leftarrow\{d\in D\mid \text{BOSS\_dist}(\delta^\star_{\mathcal{T}}(d),e)=\min_{x\in \delta^\star_E}(\text{BOSS\_dist}(\delta^\star_{\mathcal{T}}(d),x))$\; 
    
    $t_e\leftarrow\mathrm{build\_tree}(D^{+},C_e,C_b,C_r)$\;
    
    Add new branch $t_e$ to $T_B$\;
  }
}
\uElseIf{$\delta^\star$ \emph{is interval-based}}{

    \tcp{($\delta^\star_a$,$\delta^\star_v$) is the best attribute-threshold tuple to split on when $\delta^\star_\lambda$ function is applied to the interval}

    $D^\leq\leftarrow\{d\in D\mid \mathrm{get\_att\_val}\big(\delta^\star_\lambda(\langle d_{\delta^\star_s}, \cdots, d_{\delta^\star_s+\delta^\star_m-1} \rangle),\delta^\star_a\big)\leq \delta^\star_v \}$\;
    
    $t_{\mathrm{left}}\leftarrow \mathrm{build\_tree}(D^{\leq},C_e,C_b,C_r)$\;
    
    Add branch $t_{\mathrm{left}}$ to $T_B$\;
    
    $D^>\leftarrow\{d\in D\mid \mathrm{get\_att\_val}\big(\delta^\star_\lambda(\langle d_{\delta^\star_s}, \cdots, d_{\delta^\star_s+\delta^\star_m-1} \rangle),\delta^\star_a\big)> \delta^\star_v \}$\;
    
    $t_{\mathrm{right}}\leftarrow \mathrm{build\_tree}(D^{>},C_e,C_b,C_r)$\;
    
    Add branch $t_{\mathrm{right}}$ to $T_B$
}

\Return $T$
\end{algorithm}


\begin{algorithm}
\caption{$\mathrm{generate\_similarity\_splitters}(D, C_e)$}
\label{alg:gen-similarity-split}
\DontPrintSemicolon
\KwIn{$D$: a time series dataset.}
\KwIn{$C_e$: no. of similarity-based candidates}
\KwOut{$\mathcal{S}_e$: a set of similarity-based splitting functions}
\vspace*{5pt}

\tcp{Note that this algorithm is reproduced from \cite[Algorithm~2]{Lucas2018}}\;

$\mathcal{S}_e \leftarrow$ $\emptyset$ \tcp*{set of candidate similarity splitters}
\For {$i=1$ \KwTo $C_e$}{
    \tcp{sample a parameterized measure $M$ uniformly at random from $\Delta$}
    $M \xleftarrow[]{\sim}\Delta$     \tcp*{$\Delta$ is the set of 11 similarity measures used in \citep{Lucas2018}}\;
    
    \vspace*{5pt}
    \tcp{Select one exemplar per class to constitute the set $E$}
    $E \leftarrow$ $\emptyset$\;
    \ForEach{\emph{class} $c$ \emph{present in} $D$}{
      $D_c \leftarrow \left\{d\in D \mid class(d) = c\right\}$ \tcp*{$D_c$ is the data for class $c$}
      $e \xleftarrow[]{\sim}D_c$ \tcp*{sample an exemplar $e$ uniformly at random from $D_c$}
      Add $e$ to $E$
    }
    \tcp{Store measure $M$ and exemplars $E$ in the new splitter $\delta$}
    $(\delta_{M},\delta_{E}) \leftarrow (M,E)$\; 
    Add splitter $\delta$ to $\mathcal{S}_e$\;
    
}

\Return $\mathcal{S}_e$\;  
\end{algorithm}

\begin{algorithm}
\caption{$\mathrm{generate\_dictionary\_splitters}(D, C_b)$}
\label{alg:gen-dictionary-split}
\DontPrintSemicolon
\KwIn{$D$: a time series dataset}
\KwIn{$C_b$: no. of dictionary-based candidates}
\KwOut{$\mathcal{S}_b$: a set of dictionary-based splitting functions}
\vspace*{5pt}

$\mathcal{S}_b \leftarrow$ $\emptyset$ \tcp*{set of candidate dictionary splitters}

\For {$i=1$ \KwTo $C_b$}{

    \tcp{See Section~\ref{subsec:dictionary} for details of BOSS parameters}

    $\mathcal{T} \leftarrow$ select\_random\_BOSS\_transformation() 

    \vspace*{5pt}
    \tcp{Select one BOSS histogram per class to constitute the set $E$}
    $E \leftarrow$ $\emptyset$\; 

    \ForEach{\emph{class} $c$ \emph{present in} $D$}{
      
      $D_c \leftarrow \left\{d\in D \mid class(d) = c\right\}$\ \tcp*{$D_c$ is the data for class $c$} 

      $e \xleftarrow[]{\sim}D_c$\ \tcp*{sample an exemplar $e$ uniformly at random from $D_c$} 
      
      \tcp{Recall that we precomputed $\mathcal{T}(D)$ during initialization}
      Add $\mathcal{T}(e)$ to $E$ \tcp*{$\mathcal{T}(e)$ is the BOSS histogram of $e$} 
    }
    
    \tcp{Store BOSS transform $\mathcal{T}$ and exemplar histograms $E$ in the new splitter $\delta$}
    $(\delta_{\mathcal{T}},\delta_{E}) \leftarrow (\mathcal{T},E)$\;
    Add splitter $\delta$ to $\mathcal{S}_b$\;
}

\Return $\mathcal{S}_b$\;   
   
\end{algorithm}

\begin{algorithm}
\caption{$\mathrm{generate\_interval\_splitters}(D, C_r)$}
\label{alg:select-interval-split}
\DontPrintSemicolon
\KwIn{$D$: a time series dataset}
\KwIn{$C_r$: no. of interval-based candidates}
\KwOut{$\mathcal{S}_r$: a set of interval-based splitting functions}
\vspace*{5pt}

$\mathcal{S}_r \leftarrow$ $\emptyset$ \tcp*{set of candidate interval splitters}

$m_{\mathrm{min}} \leftarrow 16$ \tcp*{minimum length of random intervals}
$C^*_r \leftarrow \lfloor C_r/4 \rfloor $ \tcp*{no. of attributes per transform}
$R \leftarrow \lceil C^*_r / m_{\mathrm{min}} \rceil $ \tcp*{no. of random intervals to compute}

\vspace*{5pt}
\For {$i=1$ \KwTo $R$}{
    \tcp{Get random interval - length $m$ ($m\in[m_{\mathrm{min}},\ell]$), starting at index $s$} 
    $(\delta_{s},\delta_{m}) \leftarrow $ get\_random\_interval($m_{\mathrm{min}}, \ell$)\;
    \tcp{Add splitters for each transformation}
    \ForEach {${\delta_{\lambda}}$ \text{in} \{ACF,PACF,AR,PS \}}{
        \tcp{Apply ${\lambda}$ to each time series}
        $D_T \leftarrow\emptyset$\;
        \ForEach {$d$ in $D$}{
        \tcp{Create $d_T$, a vector of $m$ attribute-values obtained by applying $\delta_{\lambda}$ to the interval}
            $d_T \leftarrow \delta_{\lambda}( \langle d_{\delta_{s}}, \cdots d_{\delta_{s+m-1}} \rangle)$\;
            Add $d_T$ to $D_T$\;
        }
         \tcp{Calculate no.~of attributes to select from $i$th random interval and transform function $\delta_{\lambda}$}
         $A \leftarrow \lfloor C^*_r / R \rfloor $\;
        \tcp{Select at random $A$ attributes in $D_T$}
        $\tilde{P} \leftarrow $ get\_random\_attributes($D_T$, A)\;
        \ForEach {\emph{attribute} $\delta_{a}$ \emph{in} $\tilde{P}$}{
        
            $\delta_{v} \leftarrow$ find\_best\_threshold($\delta_{a}$)\;

            Add $\big( (\delta_{s},\delta_{m}), \delta_{\lambda}, (\delta_{a},\delta_{v}) \big)$ to $\mathcal{S}_r$\;

        }
    }
}

\Return $\mathcal{S}_r$\;
\end{algorithm}

\subsection{Splitting Functions}
\label{sec:splitfunctions}

As mentioned earlier, we choose splitting functions based on similarity measures, dictionary representations and interval-based transformations. This is motivated by the components of HIVE-COTE, namely EE (similarity-based), BOSS (dictionary-based) and RISE (interval-based). \reviewone{The number of \emph{candidate} splits generated per node for each type of splitter type is denoted by $C$ with a subscript as follows: $C_e$ for the number of similarity-based splitters, $C_b$ for the number of dictionary-based splitters and $C_r$ for the number of interval-based splitters.} We do not include ST (shapelets) because of its high \reviewone{training time} computational complexity. \reviewone{We also omit TSF because its accuracy is ranked lower than EE, ST and BOSS \citep{Bagnall2017}.} We next describe how we generate each of these types of splitting function.

\subsubsection{Similarity-based}
\label{subsec:similarity}
This splitting function uses the method of Proximity Forest \citep{Lucas2018}, which splits the data based on the similarity of each time series to a set of reference time series \reviewtwo{(Lines 16~to~22 in Algorithm~\ref{alg:build-tree})}. At training time, for each candidate splitter, a random measure $\delta_M$, that is randomly parameterized, is selected, as well as a set $\delta_E$ of random reference time series, one from each class (Algorithm~\ref{alg:gen-similarity-split}). \reviewone{We use the same 11 similarity measures used in Proximity Forest \citep{Lucas2018}, and the parameters for these measures are also selected randomly from the same distributions used in Proximity Forest \citep{Lucas2018}.}
If \ourmethod is trained with only the similarity-based splitter enabled (i.e. $C_b=C_r=0$), then it is exactly Proximity Forest. 

\reviewone{When designing our earlier work Proximity Forest \citep{Lucas2018} we chose to select a single random reference per class instead of an aggregate representation because it is very fast and it introduces diversity to the ensemble. We found that using a single random reference per class was working very well in Proximity Forest, and so we used it in the equivalent similarity-based splitter, and also in the dictionary-based splitter presented in Section~\ref{subsec:dictionary}.}

When splitting the data at training time and at classification time, the similarity of a query instance $Q$ to each reference time series \reviewtwo{$e$} in $\delta_E$ is evaluated using the selected measure $\delta_M$. $Q$ is passed down the branch corresponding to the \reviewtwo{$e$} to which $Q$ is closest.

\subsubsection{Dictionary-based}
\label{subsec:dictionary}
This type of split functions also uses a similarity-based splitting mechanism, except that it works on a set of time series that have been transformed using the BOSS transformation \cite[Algorithm~1]{Schafer2015}, and that it uses a variant of the Euclidean distance \cite[Definition~4]{Schafer2015} to measure  similarity between transformed time series. 

The BOSS transformation is used to convert the time series dataset into a bag-of-word model. We start by describing the BOSS transformation. To compute a BOSS transformation of a single time series, first, a window of fixed length $w$ is slid over the time series, while converting each window to a Symbolic Fourier Approximation (SFA) word of length $f$ \citep{Schafer2012, Schafer2015}. SFA is a two-step procedure: 1) it applies a low pass filter --~using only the low frequency coefficients of the Discrete Fourier Transformation (DFT)~--, 2) it converts each window (subseries) into a word using a data adaptive quantization method called Multiple Coefficient Binning (MCB). MCB defines a matrix of discretization levels for an alphabet size $\alpha$ (default is $\alpha=4$) and a word length $f$. This leads to $\alpha^f$ possible words. 
\reviewone{There is also a parameter called \emph{norm}. If it is equal to true, the first Fourier coefficient of the window is removed, which is equal to mean-centering the time series (i.e., subtracting the mean).}
SFA words are then counted to form a word frequency histogram that is used to compare two time series. BOSS uses a bespoke Euclidean distance, namely $\text{BOSS\_dist}$, \reviewone{which measures the distance between sparse vectors (which here represent histograms) in a non-symmetric way, such that the distance is computed only on elements present in the first vector \citep{Schafer2015}.}

We now turn to explaining how we use BOSS transformations to build our forest. Since BOSS has four different hyperparameters, many possible BOSS transformations of a time series can be generated. Before we start training the trees, $t$ BOSS transformations (histograms for all time series) of the dataset are pre-computed based on $t$ randomly selected sets of BOSS parameters. Similar to the values used in BOSS, the four parameters are selected uniformly at random from the following ranges: the window length $w \in \{10 \cdots \ell\}$, SFA word length $f \in \{6,8,10,12,14,16\}$, the normalization parameter $norm \in \{true, false\}$, and $\alpha=4$. 

At training time (Algorithm~\ref{alg:gen-dictionary-split}), for each candidate splitter $\delta$, a random BOSS transformation $\delta_\mathcal{T}$\reviewone{, with replacement,} is chosen, as well as a set $\delta_E$ of random reference time series from each class for which the transformation $\delta_\mathcal{T}$ has been applied. Each training time series is then passed down the branch of the reference series for which the BOSS distance between histogram of the series and the reference time series is lowest. We then generate several such splitters and choose the best one according to the Gini index.

At classification time, when a query time series $Q$ arrives at a node with a dictionary-based splitter, we start by calculating its transformation into a word histogram (the transformation $\delta_\mathcal{T}$ selected at training). We then compare this histogram to each reference time series in $\delta_E$, and $Q$ is passed down the branch corresponding to the reference time series to which $Q$ is closest.

\subsubsection{Interval-based}
\label{subsec:interval}
This type of splitting function is designed to work in a similar fashion to the RISE component used in the HIVE-COTE. Recall that RISE is an interval-based algorithm that uses four transformations (ACF, PACF, AR - in time domain and PS - in frequency domain) to convert a set of random intervals to a feature vector. Once the feature vectors have been generated, RISE uses a classic attribute-value splitting mechanism to train a forest of binary decision trees (similar to Random Forest\reviewone{ -- but without bagging}). 

\reviewone{A notable difference between RISE, and our interval-based splitter is that the random} intervals are selected per tree in \reviewone{RISE}, whereas \reviewone{our interval-based splitter} selects random intervals per candidate split at the node level. 
\reviewone{This choice is for two main reasons. Firstly, choosing intervals per candidate split at node level helps to explore a larger number of random intervals. Secondly, this also separates the hyperparameter $k$ (number of trees) from the number of random intervals used by the interval-based splitters which depends on the hyperparameter $C_r$ (number of interval-based splits per node). Separating these hyperparameters helps to change the effects of interval-based splitter on the overall ensemble, without changing the size of the whole ensemble.
Consequently, this design decision also helps to increase the diversity of the ensemble.}

\reviewone{Algorithm~\ref{alg:select-interval-split} describes the process of generating features using random intervals and the four transform functions to generate $C_r$ interval-based candidates splits.}
Each candidate splitter $\delta$ is defined by a pair $(\delta_s,\delta_m)$ that represent the interval start and its length respectively, a function $\delta_\lambda$ (one of ACF, PACF, AR or PS) which is applied to the interval and a pair $(\delta_a,\delta_v)$ that indicates the attribute $\delta_a$ and threshold value $\delta_v$ on which to split. The values of $(\delta_s,\delta_m)$ are \reviewone{randomly selected to get a random interval of length between minimum length $m_{\mathrm{min}}=16$ and $\ell$ the length of the time series. We set $m_{\mathrm{min}}$, and other parameters required by the four transform functions to be exactly same as it was in RISE.} The values of the pair $(\delta_a,\delta_v)$ are optimized such that the Gini index is maximized when the data are split on the attribute $\delta_a$ for a threshold value $\delta_v$.

When splitting the data at training time and at classification time, $\delta_{\lambda}$ is applied to the interval of query instance $Q$ defined by $\delta_s$ and $\delta_m$, obtaining the attribute vector $Q_\lambda$. If 
$\mathrm{get\_att\_val} (Q_{\lambda},\delta_a)\leq\delta_v$ (the value of attribute ${\delta_a}$ of ${Q_{\lambda}}$ is less than the threshold value), $Q$ is passed down the left branch.  Otherwise it is passed down the right. 
\reviewone{Contrary to the similarity- and dictionary-based splitting functions, which used a distance based mechanism to partition the data (to produce a variable number of branches depending on the number of classes present at the node), the ``attribute-value'' based splitting mechanism used by the interval-based splitting functions produce binary splits \reviewtwo{(Lines 32~to~38} in Algorithm~\ref{alg:build-tree}).}

\subsection{Classification}

For each tree, a query time series $Q$ is passed down the hierarchy from the root to the leaves. The branch taken at each node depends on the  splitting function selected at the node. Once $Q$ reaches the leaf, it is labelled with the class with which the training instances that reached that leaf were classified. Recall that the tree is repeatedly split until pure, so all training instances that reach a leaf will have the same class.  This process is presented in the Algorithm~\ref{alg:classification}. Finally, a majority vote by the $k$ trees is used to label $Q$.

\begin{algorithm}
\caption{classification$\left(Q, T\right)$}
\label{alg:classification}
\DontPrintSemicolon
\KwIn{$Q$: Query Time Series}
\KwIn{$T$: \ourmethod Tree}
\KwOut{a class label $c$}
\uIf{$\mathrm{is\_leaf}(T)$}{\Return majority class of $T$}

 \uIf{$T_\delta$ \emph{is similarity-based}}{
   $(e, T^\star) \leftarrow \argmin\limits_{(e',T')\in T_B} \delta_M(Q,e')$\;
  }
  \uElseIf{$T_\delta$ \emph{is dictionary-based}}{
    $(e, T^\star) \leftarrow \argmin\limits_{(e',T')\in T_B} \text{BOSS\_dist}(\delta_\mathcal{T}(Q),e')$\; 
 }
 \uElseIf{$T_\delta$ \emph{is interval-based}}{
   $Q_\lambda \leftarrow \delta_\lambda(\langle Q_{\delta_s}, \cdots, Q_{\delta_s+\delta_m-1} \rangle)$\;
   \tcp{compare the $\delta_a^{th}$ attribute value from $Q_\lambda$ to the split value}
   \uIf{$\mathrm{get\_att\_val}(Q_{\lambda},\delta_a) \le \delta_v$}{
       $T^* \leftarrow T_{left}$
   }
   \uElse{
        $T^* \leftarrow T_{right}$
   }
   }

\tcp{recursive call on subtree $T^\star$}
\Return classification$\left(Q, T^\star\right)$\; \end{algorithm}

\subsection{Complexity}
\label{subsec:complexity}

\paragraph{Training time complexity}
Proximity Forest, on which TS-CHIEF builds, has average training time complexity that is quasi-linear with the quantity of training data, $O(k \cdot n\log(n) \cdot C_e \cdot c \cdot \ell^2)$ for $k$ trees, $n$ training time series of length $\ell$, $C_e$ similarity-based candidate splits, and $c$ classes \citep{Lucas2018}. The term $k$ comes from the number of trees to train and $\log(n)$ from the average depth of the trees. In the worst case, tree depth may be $n$, however, on average, tree depth can be expected to be $\log(n)$. The term $n \cdot C_e \cdot c \cdot \ell^2$ represents the order of time required to select the best of $C_e$ candidate splits and partition the data thereon, based on the similarity of $n$ training instances to $c$ reference time series at the node using a random similarity measure. The slowest of the similarity measures used (WDTW) is bounded by $O(\ell^2)$.

The addition of the dictionary-based splitter adds a new  initialization step and a new selection step to the Proximity Forest algorithm. The initialization part pre-computes $t$ BOSS transformations for $n$ time series. Since the cost of BOSS transforming one time series is $O(\ell)$ \cite[Section~6]{Schafer2015}, the complexity of the initialization part is $O(t \cdot n \cdot \ell)$. The Euclidean-based BOSS distance has a complexity of $O(\ell)$ \cite[Definition~4]{Schafer2015} and must be applied to every example at the node for each of the $C_b$ (dictionary-based candidate splits), resulting in order $O(C_b\cdot c \cdot n \cdot \ell)$ complexity for generating and evaluating dictionary splitters at each node of each tree.

 The interval-based splitting functions are attribute-value splitters; we detail the complexity for training a node receiving $n'$ time series. Each interval is transformed using 4 different functions (ACF, PACF, AR and PS), which takes at most $O(\ell^2)$ time \cite[Table~4]{Lines2018}, leading to $O(r\cdot n'\cdot \ell^2)$ for $r$ intervals taken where $r$ is proportional to $C_r$.
 For each of the $C_r$ candidate splits the data is then sorted and scanned through to find the best split --~$O(C_r\cdot n\log(n))$. Put together, this adds $O(C_r\cdot n\cdot \ell^2 + C_r\cdot n\log(n))$ complexity to the split selection stage. Note that $\ell$ in this term represents an upper bound on the length of random intervals selected. The expected length of random interval is 1/3 of $\ell$.

 Overall, \ourmethod has quasi-linear average complexity \reviewone{with respect to the training size} : 
 \begin{align*}
 O\Big(\underbrace{t\cdot n\cdot \ell}_\text{initialization} + \underbrace{k \cdot \log(n)}_{\substack{\text{avg.depth} \\ \text{for $k$ trees}}} \thinspace \cdot \thinspace
 \big[&\underbrace{C_e \cdot c \cdot n\cdot \ell^2}_\text{similarity} 
 + \underbrace{C_b\cdot c\cdot n  \cdot \ell}_\text{dictionary} \\
 &+ \underbrace{C_r\cdot n\cdot \ell^2 
 + C_r\cdot n\log(n)}_\text{interval}\big]\Big).
 \end{align*}

In Section~\ref{subsec:accuracy_ablative}, we have included an experiment to measure the fraction of training time taken by each splitter type over 85 UCR datasets \citep{UCRArchive2015}. As expected, the dominant term in the training complexity is the term representing the similarity-based splitter. 
In practice, our experiments show that the similarity-based splitter takes about $80\%$ of the training time (See Figure~\ref{fig:splitter-timing}, on page \pageref{fig:scale-sat-time}).

\paragraph{Classification time complexity}
Each time series is simply passed down $k$ trees, traversing an average of $log(n)$ nodes. Moreover, the complexity at each node is dominated by the similarity-based splitters. Overall, this is thus a $O(k\cdot \log(n)\cdot c\cdot \ell^2)$ average case classification time complexity. 
 
\paragraph{Memory complexity}
The memory complexity is linear with the quantity of data. We would need to store one copy of $n$ time series of length $\ell$ -- this is $O(n \cdot \ell)$. In the worst case there are as many nodes in each of the $k$ trees as there are time series and at each node, and we store one exemplar time series for each of the $c$ classes, $O(k\cdot n\cdot c)$. We pre-store all $t$ dictionary-based transformations, $O(t\cdot n \cdot \ell)$. Overall, this is $O(n \cdot \ell +  k\cdot n\cdot c + t\cdot n \cdot \ell)$.

\section{Experiments}
\label{sec:experiments}
We start by evaluating the accuracy of \ourmethod on the UCR archive, and then assess its scalability on a large time series dataset. In essence, we show that \ourmethod can reach the same level of accuracy as HIVE-COTE but with much greater speed, thanks to \ourmethod's quasi-linear complexity \reviewtwo{with respect to the} \reviewone{number of training instances}.
\reviewone{We then present a study on the variation of training accuracy against the ensemble size, followed by an assessment of the contribution of each type of splitter in \ourmethod. Finally, we finish this section by presenting a study of the memory requirements for \ourmethod.}

\reviewone{We implemented a multi-threaded version of \ourmethod in Java, and have made it available via the Github repository \url{\gitrepo}. In these experiments, we used multiple threads when measuring the accuracy of \ourmethod under various configurations (Sections \ref{subsec:accuracy_ucr}, \ref{subsec:accuracy_var} and \ref{subsec:accuracy_ablative}). However, we used a single thread (1~CPU) for both \ourmethod and HIVE-COTE when measuring the timings for scalability experiments in Section \ref{subsec:scalability}.}

Throughout the experiments, unless mentioned otherwise, we use the following parameter values for \ourmethod: $t=1000$ dictionary-based (BOSS) transformations, $k=500$ trees in the forest. When training each node, we concurrently assess the following number of candidates: 5 similarity-based splitters, 100 dictionary-based splitters and 100 interval-based splitters. Ideally, we would also want to raise the number of candidates for the similarity-based splitter, but this has a significant impact on training time (since passing the instances down the branches measures in $O(\ell^2)$) with marginal improvement in accuracy \citep{Lucas2018}. 
Note that we have not done any tuning of these numbers of candidates of each type. \reviewone{For hyperparameters of the similarity-based splitters (e.g. parameters for distance measures), we used exactly the same values used in Proximity Forest \citep{Lucas2018}. Similarly, for dictionary- and interval-based splitters, we used the same hyperparameters used in BOSS and RISE components of HIVE-COTE \citep{Lines2018}.}

\subsection{Accuracy on the UCR Archive}
\label{subsec:accuracy_ucr}
We evaluate \ourmethod on the UCR archive \citep{UCRArchive2015}, as is the de facto standard in TSC research \citep{Bagnall2017}. We use the 2015 version with 85 datasets, because the very recent update adding further datasets is still in beta \citep{Dau2018a}. All 85 datasets are fixed length univariate time series that have been $z$-normalized. We use the standard train/test split available at \url{http://www.timeseriesclassification.com}. 

\reviewone{
To compare multiple algorithms over the 85 datasets, we use critical difference diagrams, as it is the standard in machine learning research \citep{Demsar2006, Benavoli2016}. We~use the Friedman test to compare the ranks of multiple classifiers \citep{Demsar2006}. In these statistical tests, the null hypothesis corresponds to no significant difference in the mean rankings of the multiple classifiers (at a statistical significant level $\alpha=0.05$). In cases where null-hypothesis was rejected, we use the Wilcoxon signed rank test to compare the pair-wise difference in ranks between classifiers, while using Holm's correction to adjust for family-wise errors \citep{Benavoli2016}.
}

We compare \ourmethod to the 3 time series classifiers identified by \citep{Bagnall2017} as the most accurate on the UCR archive (FLAT-COTE, ST and BOSS), as well as the de facto standard 1-NN DTW, deep learning method ResNet and the more recent HIVE-COTE (the current most accurate on the URC archive) and Proximity Forest (the inspiration for TS-CHIEF).  We use results reported at the \url{http://www.timeseriesclassification.com} website for these algorithms, except for TS-CHIEF, Proximity Forest (our result \citep{Lucas2018}) and the deep learning ResNet method for which we obtained the results from Fawaz et. al's review of Deep Learning methods for TSC \citep{Fawaz2018}.

\begin{figure}[!ht]
    \centering
    \includegraphics[width=.8\linewidth]{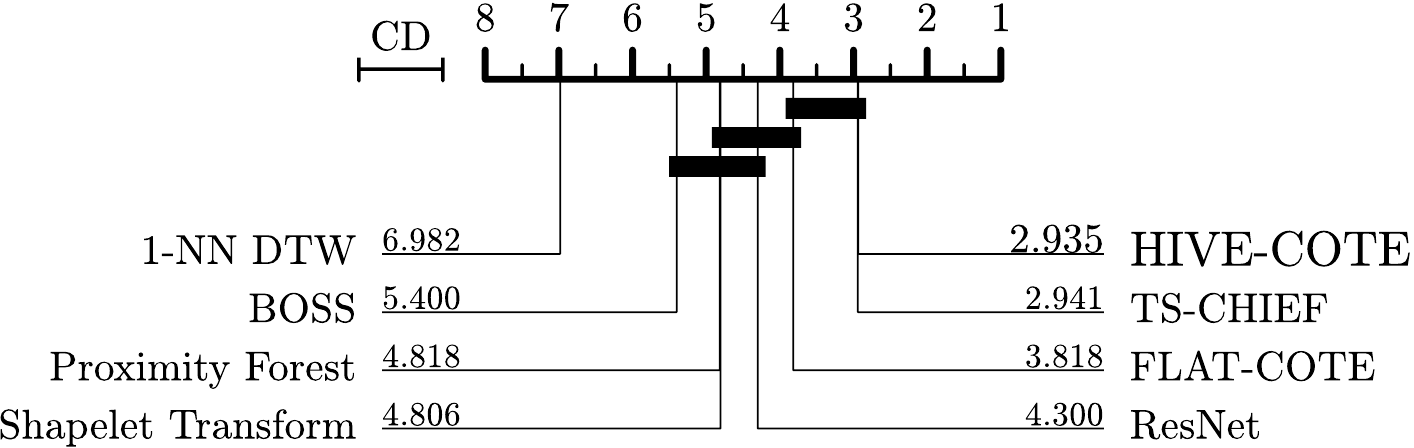}
    \caption{Critical difference diagram showing the average ranks on error of leading TSC algorithms (described in Section \ref{sec:relatedwork}) across 85  datasets from the benchmark UCR archive \citep{Dau2018a}.
    The lower the rank (further to the right) the lower the error of an algorithm relative to the others on average.}
    \label{fig:cd-us-vs-tsc}
\end{figure}

\reviewone{
Figure~\ref{fig:cd-us-vs-tsc} displays mean ranks (on error) between the 8 algorithms; which is also the main result of this paper in terms of accuracy. \ourmethod obtains an average rank of 2.941, which rivals HIVE-COTE at 2.935 (statistically not different). FLAT-COTE comes next with an average rank of 3.818. Next, Residual Neural Network (ResNet) is ranked at 4.300. 

Table~\ref{tbl:crit} presents the results of a  comparison between each pair of algorithms. We use Wilcoxon's signed rank test and judge significance at the 0.05 significance level using a Holm correction for multiple testing. The comparisons that are judged significant at the 0.05 level are displayed in bold type. \ourmethod, HIVE-COTE, FLAT-COTE and ResNet are all statistically indistinguishable from one another except that HIVE-COTE is significantly more accurate than FLAT-COTE. \ourmethod and the two COTEs are all significantly more accurate than all the other algorithms except ResNet.}

\begin{table}[h!]
\centering
    \begin{tabular}{lrrrrrrrr}
    \toprule
    {}  &  \textbf{BOSS} &    \textbf{ST} &    \textbf{PF} &  \textbf{RN} &  \textbf{FCT} &  \textbf{HCT} &  \textbf{\ourmethod} \\
    \midrule

    \textbf{DTW} & \textbf{$<$0.001} & \textbf{$<$0.001} & \textbf{$<$0.001} & \textbf{$<$0.001} & \textbf{$<$0.001} & \textbf{$<$0.001} & \textbf{$<$0.001} \\
    \textbf{BOSS} &       & 0.035 & 0.042 & 0.022 & \textbf{$<$0.001} & \textbf{$<$0.001} & \textbf{$<$0.001} \\
    \textbf{ST} &       &       & 0.684 & 0.112 & \textbf{$<$0.001} & \textbf{$<$0.001} & \textbf{$<$0.001} \\
    \textbf{PF} &       &       &       & 0.127 & \textbf{0.002} & \textbf{$<$0.001} & \textbf{$<$0.001} \\
    \textbf{RN} &       &       &       &       & 0.330 & 0.005 & 0.017 \\
    \textbf{FCT} &       &       &       &       &       & \textbf{$<$0.001} & 0.045 \\
    \textbf{HCT} &       &       &       &       &       &       & 0.687 \\

    \bottomrule
    \end{tabular}
\caption{\reviewone{p-values for the pairwise comparison of classifiers. Bold values indicate pairs of classifiers that are statistically different at the 0.05 level after applying a Holm correction. The algorithms are abbreviated as follows. RN: ResNet, DTW: 1-NN DTW, FCT: FLAT-COTE, and HCT: HIVE-COTE.}} 

\label{tbl:crit}
\end{table}

\begin{figure}[!ht]
     \centering
     \begin{subfigure}[b]{.45\textwidth}
         \centering
         \includegraphics[width=\textwidth]{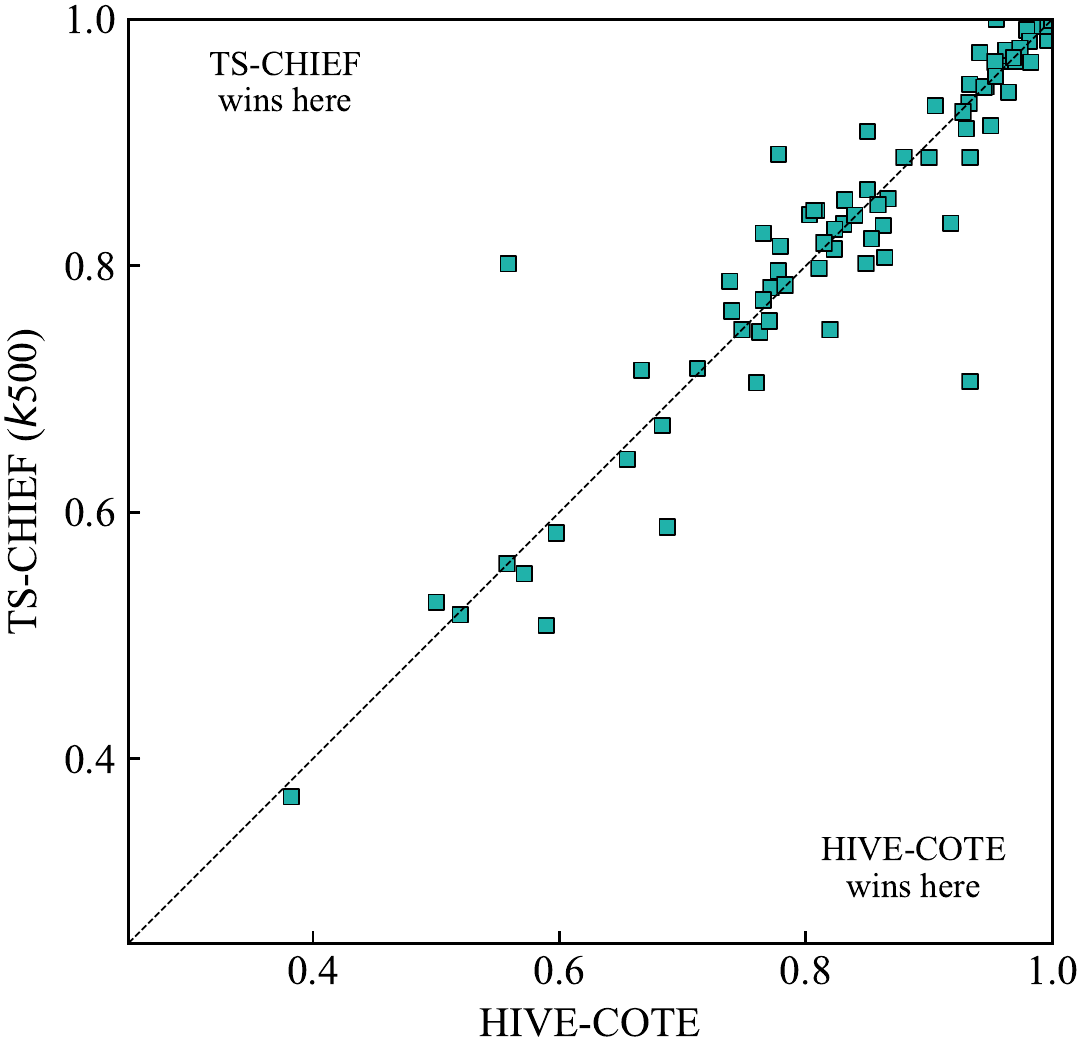}
         \label{fig:acc-best-vs-hcote-a}
     \end{subfigure}
     \begin{subfigure}[b]{0.45\textwidth}
         \centering
         \includegraphics[width=\textwidth]{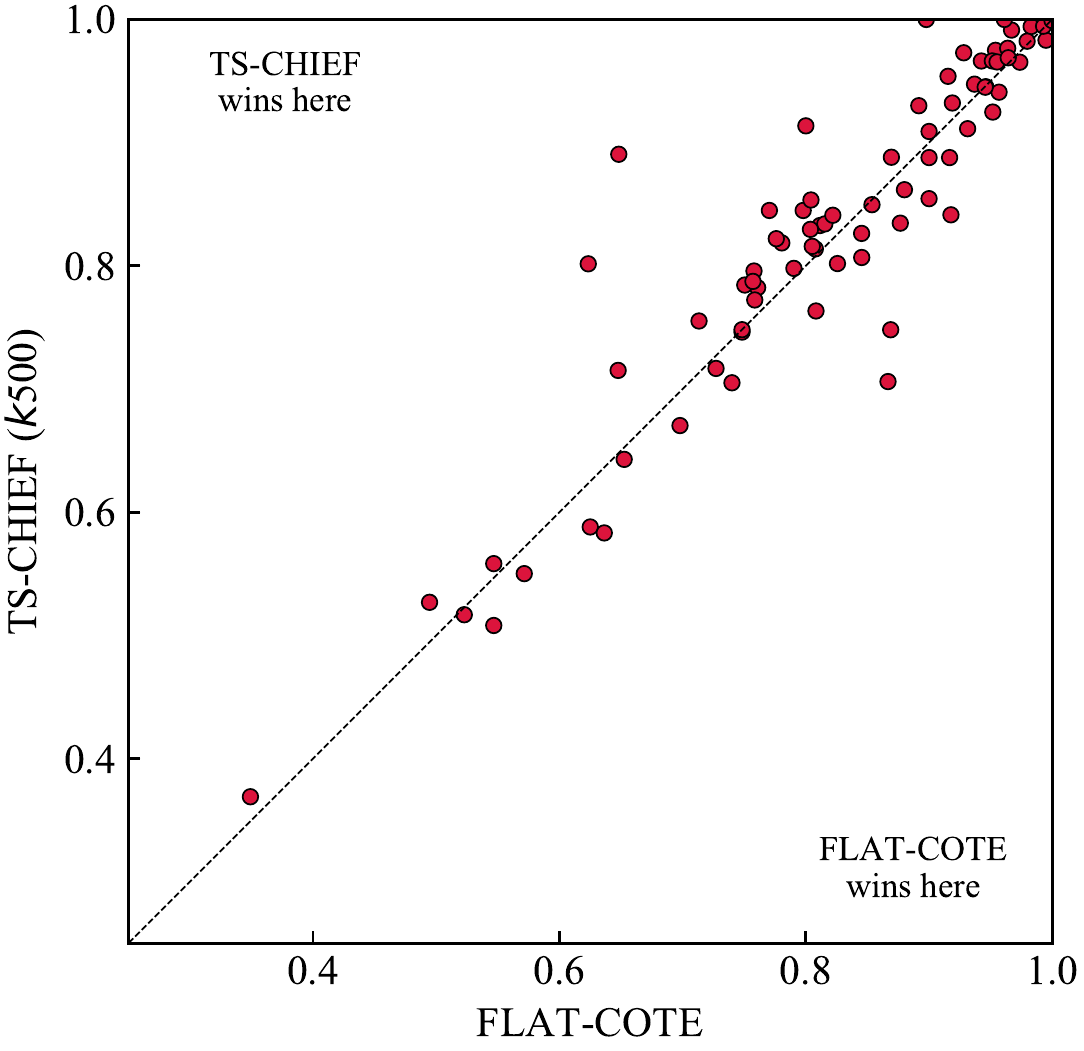}
         \label{fig:acc-best-vs-hcote-b}
     \end{subfigure}
     \vspace*{-8pt}
    \caption{\reviewone{Comparison of accuracy for \ourmethod \emph{versus} HIVE-COTE (left) and \ourmethod \emph{versus} FLAT-COTE (right) on 85 UCR datasets. \ourmethod's win/draw/loss against  HIVE-COTE is 40/7/38 and against FLAT-COTE is 47/5/33.}}
\label{fig:acc-best-vs-hcote}
\end{figure}

To further examine the accuracy of \ourmethod against both COTE algorithms, Figure~\ref{fig:acc-best-vs-hcote} presents a scatter plot of pairwise accuracy.
Each point represents a UCR dataset. \ourmethod wins above the diagonal line. \ourmethod wins 40 times against HIVE-COTE (green squares), loses 38 times and ties on 7 datasets. Compared to FLAT-COTE (red circles), \ourmethod wins 47 times, and loses 33 times, with 5 ties. It is interesting to see that \ourmethod  gives results that are quite different to both COTE algorithms, with a few datasets for which the difference in accuracy is quite large.

\reviewone{Table~\ref{tbl:results} (on page \pageref{tbl:results}) presents the accuracy of all 8 classifiers for the 85 datasets. \ourmethod is most accurate of all classifiers (rank 1) on 31 datasets, while HIVE-COTE is most accurate on 23, despite their mean ranks being equal at 2.94. With respect to the benchmark UCR archive, \ourmethod rivals HIVE-COTE in accuracy (without being statistically different).}

\begin{table}
\centering
    \begin{tabular}{lrrrrrrrr}
    \toprule
    {} &   DTW &  BOSS &    ST &    PF &  RN &  FCT &  HCT &  CHIEF \\
    Dataset Type &       &       &       &       &         &        &        &        \\
    \midrule
    DEVICE       & 59.54 & 66.81 & 70.58 & 64.40 &   72.94 &  69.47 &  \textbf{73.24} &  69.26 \\
    ECG          & 87.14 & 91.69 & 94.43 & 92.34 &   92.87 &  \textbf{95.56} &  95.20 &  94.88 \\
    IMAGE        & 74.87 & 81.27 & 79.71 & 82.30 &   79.79 &  82.67 &  84.05 &  \textbf{84.35} \\
    MOTION       & 70.54 & 75.60 & 77.87 & 78.55 &   76.83 &  79.13 &  79.66 &  \textbf{81.40} \\
    SENSOR       & 77.50 & 79.89 & 84.05 & 83.66 &   85.77 &  \textbf{86.01} &  84.81 &  84.67 \\
    SIMULATED    & 87.25 & 92.61 & 92.20 & 89.26 &   93.13 &  93.96 &  94.49 &  \textbf{94.79} \\
    SPECTRO      & 80.34 & 85.00 & 86.55 & 81.67 &   86.19 &  84.72 &  \textbf{88.31} &  86.49 \\
    \bottomrule
    \end{tabular}
\caption{\reviewone{Mean accuracy of TSC algorithms grouped by dataset types identified in the UCR archive \citep{UCRArchive2015}. FCT and HCT indicates FLAT-COTE and HIVE-COTE respectively, and RN indicates ResNet.}}
\label{tbl:dataset-type}
\end{table}

\reviewone{
We also looked at the accuracy of \ourmethod and other TSC methods on different data domains as identified in the UCR achive \citep{UCRArchive2015}. The results, in Table~\ref{tbl:dataset-type}, shows that \ourmethod performed best in three data domains, although the mean accuracy in these cases are similar to HIVE-COTE.
}

Although we were not able to compare running time with either of the COTE algorithms because of their very high running time, even on the UCR archive, we give here a few indications of runtime for \ourmethod. The experiment was carried out using an AMD Opteron CPU (1.8~GHz) with 64~GB RAM, with 16~CPU threads. \reviewone{Note that this is the only timing experiment we ran with multiple threads, timing experiments in Section~\ref{subsec:scalability} were run using a single thread.}

Average training and testing times were respectively of about 3 hours and 27 min per dataset, but with quite a large difference between datasets. \ourmethod was trained on 69 datasets in less than 1 hour each and less than one day was sufficient to train \ourmethod on all but 10 datasets. It however took about 10 days to complete training on all the datasets, mostly due to the \texttt{HandOutlines} dataset which took more than 4 days to complete. Our experiments confirmed our theoretical developments about complexity: \ourmethod was largely unaffected by dataset size with the largest dataset \texttt{ElectricDevices} trained in 2h24min and tested in 9min. \texttt{HandOutlines} is the dataset with the longest series and in the top-10 in terms of training size, which shows that the quadratic complexity with the length has still a non-negligible influence on training time. 
The next section details scalability \reviewtwo{with respect to} length and size.

\subsection{Scalability}
\label{subsec:scalability}
\ourmethod is designed to be both accurate and highly scalable. Section~\ref{subsec:complexity} showed that the complexity of \ourmethod scales quasi-linearly \reviewtwo{with respect to} number of training instances $n$ and quadratically \reviewtwo{with respect to} length of the time series $\ell$. To assess how this plays out in practice, we carried out two experiments to evaluate the runtime of \ourmethod when 1)~the number of training instances increases, and 2)~the time series length increases. We compare \ourmethod to the HIVE-COTE algorithm which previously held the title of most accurate on the UCR archive. We performed these experiments with 100 trees.
As the accuracy on the UCR archive has been evaluated for 500 trees (Section~\ref{subsec:accuracy_ucr}), we also estimated the timing for 500 trees (5 times slower). The experiments used a single run of each algorithm using 1~CPU \reviewone{(single thread)} on a machine with an Intel(R) Xeon(R) CPU E5-2680 v3 @ 2.50GHz processor with 200 GB of RAM.  

\subsubsection{Increasing training set size}
First, we assessed the scalability of \ourmethod \reviewtwo{ with respect to the} training set size. We used a Satellite Image Time Series (SITS) dataset \citep{Tan2017} composed of 1 million time series of length 46, with 24 classes. \reviewone{The training set was sampled using stratified random sampling method while making sure at least one time series from each class in the training data is present in the stratified samples. We~also used a stratified random sample of 1000 test instances for evaluation.} We~evaluated the accuracy and the total runtime as a function of the number of training time series, starting from a subsample of 58, and logarithmically increasing up to 131,879 (a sufficient quantity to clearly define the trend). 

\begin{figure}[!ht]
    \centering
    \includegraphics[width=0.6\linewidth]{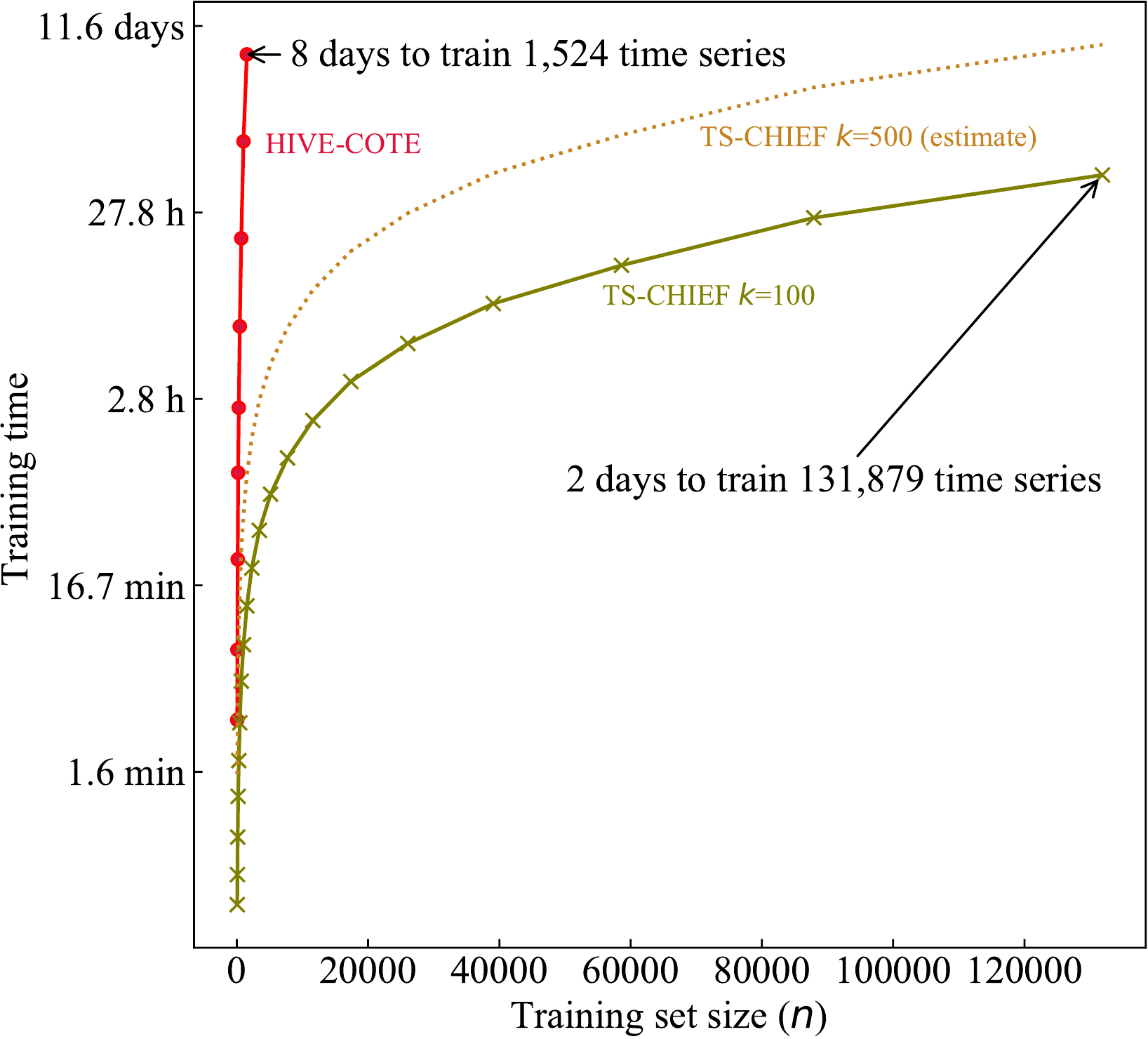}
    \vspace*{-8pt}
    \caption{Training time in logarithmic-scale for \ourmethod \emph{versus} HIVE-COTE with increasing training size using the Satellite Image Time Series dataset \citep{Tan2017}. Even for 1,500 time series, \ourmethod is 
    more than 900 times faster than the current state of the art HIVE-COTE.}
    \label{fig:scale-sat-time}
\end{figure}

\begin{figure}[!htb]
\centering
\includegraphics[width=.6\linewidth]{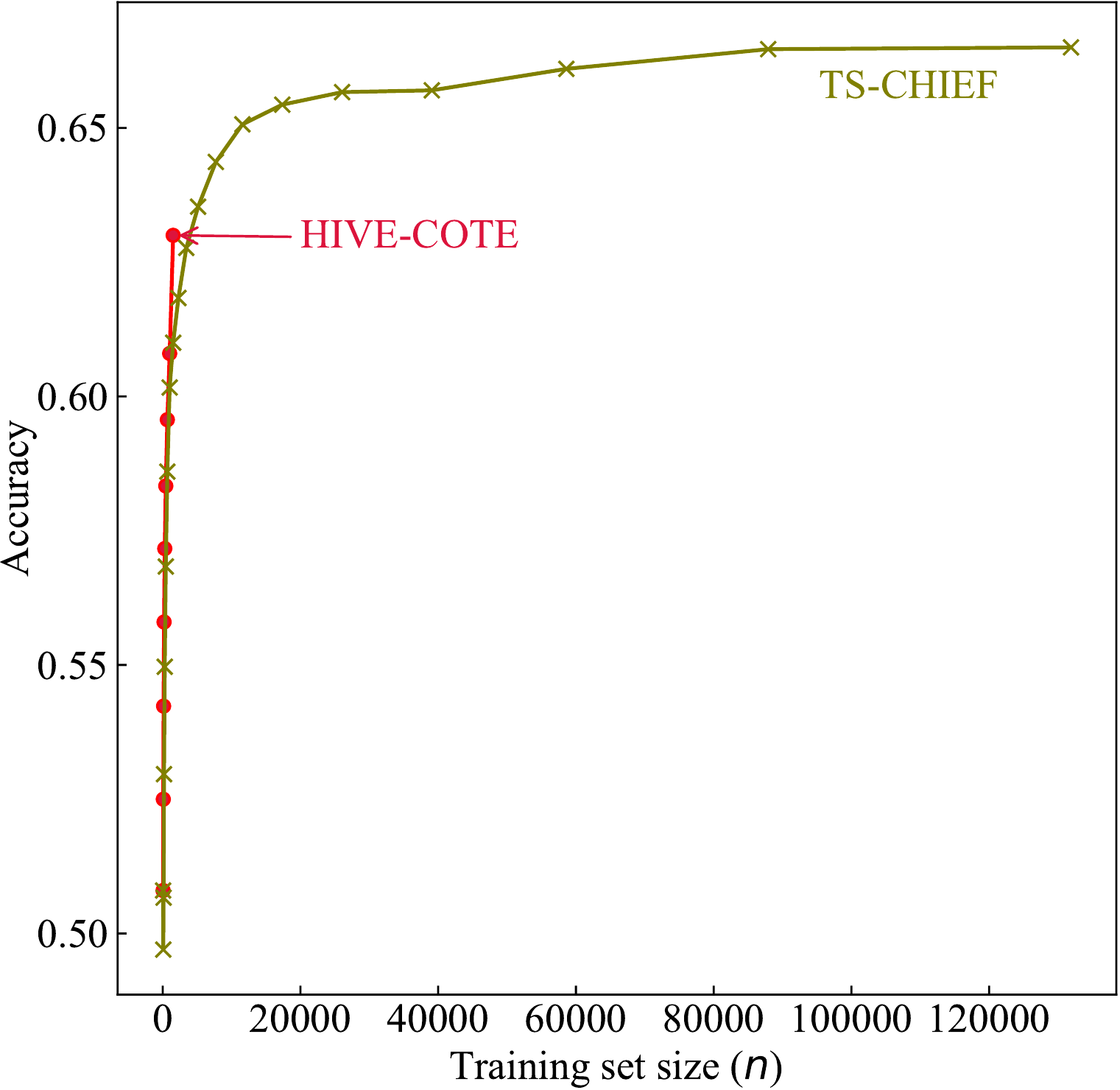}
\vspace*{-8pt}
\caption{Accuracy as a function of training set size for SITS dataset.}
\label{fig:scale-sat-acc}
\end{figure}

Figures~\ref{fig:scale-sat-time} and \ref{fig:scale-sat-acc} show the training time and the accuracy, respectively, as a function of the training set size for \ourmethod (in olive) and HIVE-COTE (in red). Figure~\ref{fig:scale-sat-time} shows that \ourmethod trains in time that is quasi-linear \reviewtwo{with respect to} the number of training examples, rather than the quadratic time for HIVE-COTE. For about 1,500 training time series, HIVE-COTE requires about 8 days to train, while \ourmethod was able to train in about 13 minutes. This is thus an 900x speed-up.

Figure~\ref{fig:scale-sat-acc} shows  that  \ourmethod has similar accuracy to HIVE-COTE  for any given number of training time series. However,  \ourmethod achieves 67~\% accuracy within 2 days by learning from about 132k time series. By fitting a quadratic curve through HIVE-COTE training time, we estimate that it will require 234 years for HIVE-COTE to learn from 132k time series. This is a speed-up of 46,000 times over HIVE-COTE. Furthermore, to train all one million time series in the SITS dataset, we estimated that it would take 13,550 years to train HIVE-COTE, while \ourmethod is estimated to take 44 days. This is a speed-up of 90,000 times over HIVE-COTE for 1M time series.

Moreover, Figure~\ref{fig:scale-sat-acc} indicates that HIVE-COTE can only achieve 60~\% after 2 days of training, i.e. a decrease of 7.9~\% compared to \ourmethod.
In practice, the execution time of \ourmethod thus scales very close to its theoretical average complexity (Section~\ref{subsec:complexity}) by scaling quasi-linearly with the training set size.

\subsubsection{Increasing length}
Second, we assessed the scalability of \ourmethod \reviewtwo{ with respect} to the length $\ell$ of the time series. We use here \texttt{InlineSkate}, a UCR dataset composed of 100 time series and 550 test time series of original length 1882. We resampled the length from 32 to 2048 by using an exponential scale with base~2.

\begin{figure}[!htb]
\centering
\includegraphics[width=.6\linewidth]{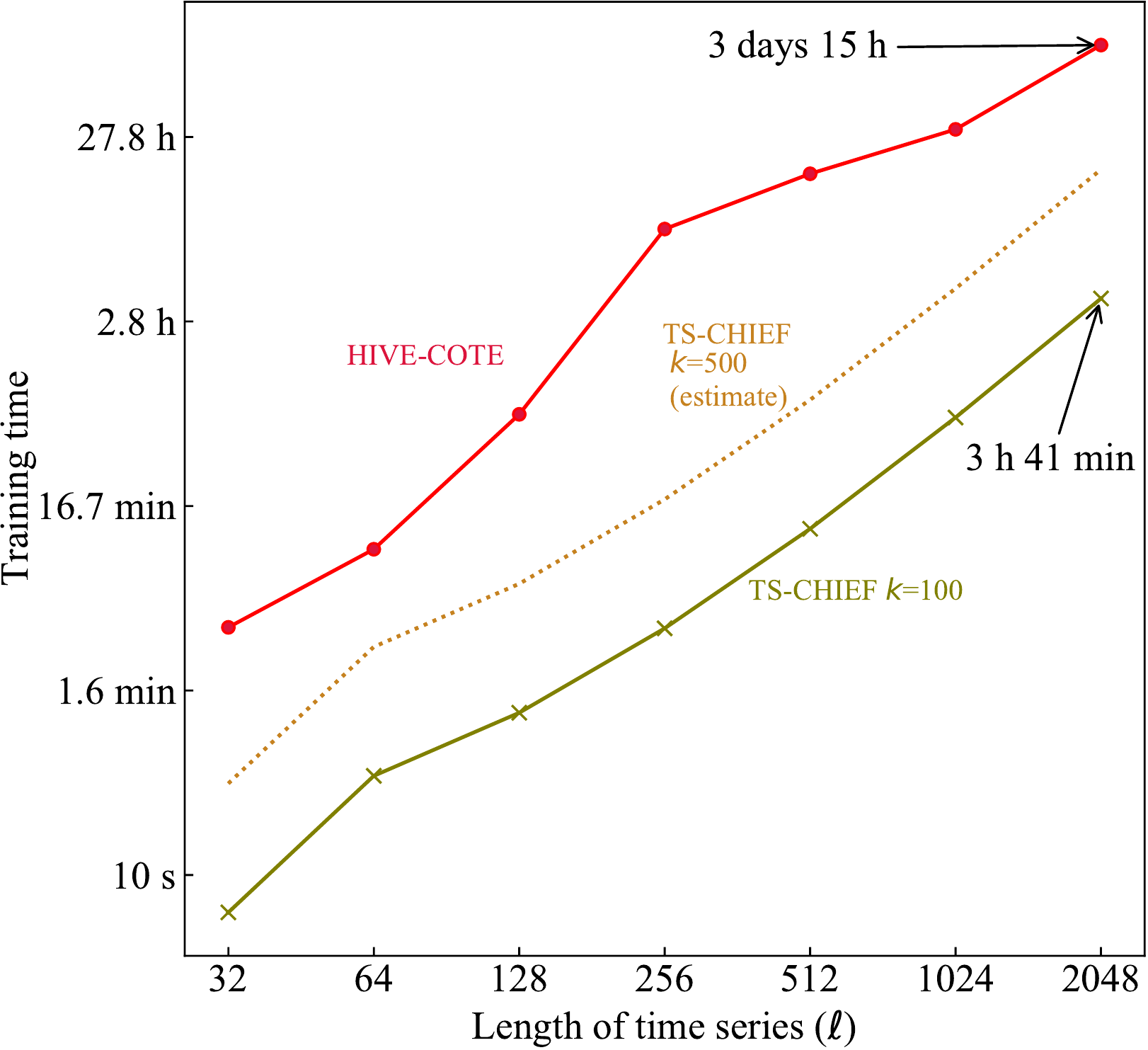}
\vspace*{-8pt}
\caption{Training time as a function of the series length $\ell$ for a one UCR dataset.}
\label{fig:scale-length-time}
\end{figure}

Figure~\ref{fig:scale-length-time} displays the training time for both \ourmethod (in olive) and HIVE-COTE (in red) as a function of the length of the time series. \ourmethod can learn from 100 time series of length 2,048 in about 4 hours, while HIVE-COTE requires more than 3 days. This is a 24x speed up. It also mirrors the theoretical training complexity of \ourmethod in $O(\ell^2)$, and HIVE-COTE in $O(\ell^4)$ \citep{Lines2018}  \reviewtwo{with respect to the} length of the time series.

\reviewone{
    \subsection{Ensemble Size and Variance of the Results}
    \label{subsec:accuracy_var}

    \begin{figure}[!htb]
         \centering
         \begin{subfigure}[b]{.6\textwidth}
             \centering
             \includegraphics[width=\textwidth,height=6.5cm]{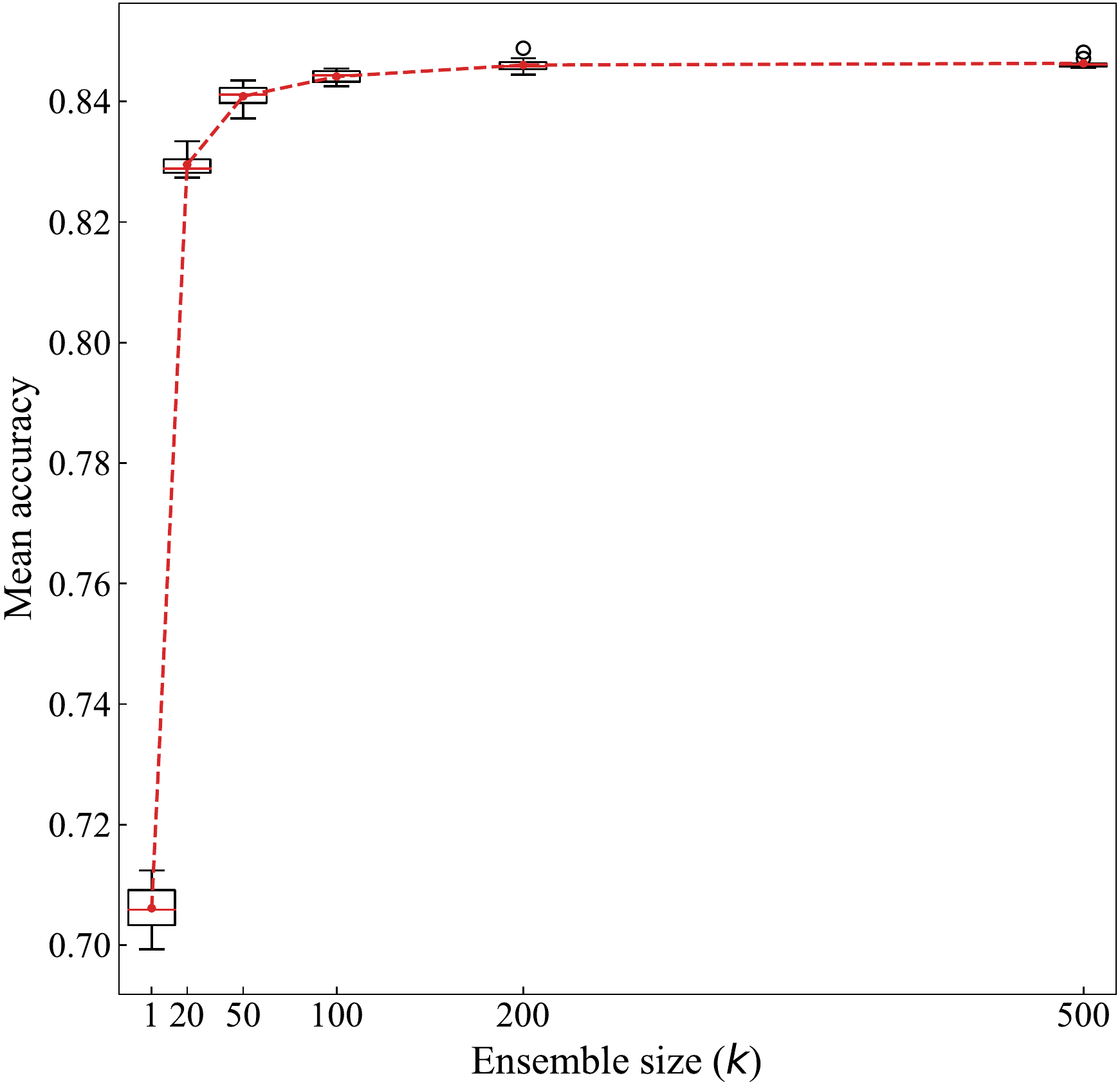}
             \label{fig:acc-var-a}
         \end{subfigure}
         \\ 
         \begin{subfigure}[b]{0.8\textwidth}
             \centering
             \includegraphics[width=\textwidth]{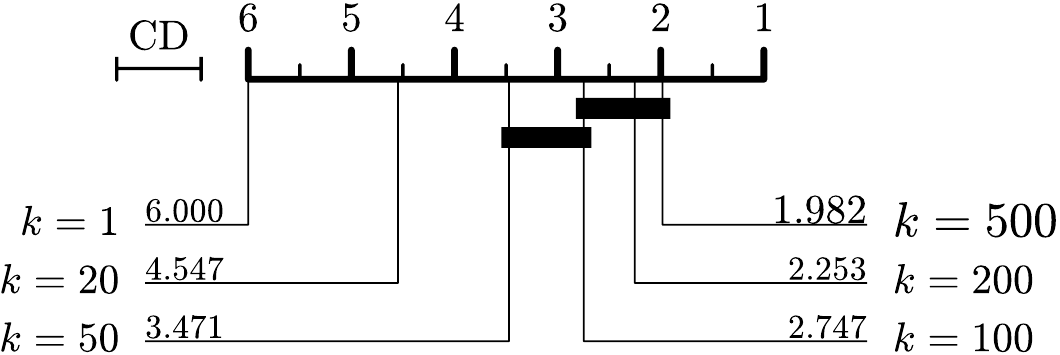}
             \label{fig:acc-var-b}
         \end{subfigure}
         \vspace*{-8pt}
         \caption{\reviewone{Mean accuracy (and variance) \emph{versus} ensemble size (top) and a critical difference diagram showing the mean rankings of different ensemble sizes (bottom).
         Mean accuracy is calculated over 85 datasets for 10 runs.}}
    \label{fig:acc-var}
    \end{figure}

    We also conducted an experiment to study the accuracy (and variance) \emph{versus} ensemble size $k$ (see Figure~\ref{fig:acc-var}). It shows that the accuracy increases with $k$ up to a point where it plateaus. This follows ensemble theory which shows that increasing the size of the ensemble reduces the variance, but that at some point this variance is compensated by the covariance of the elements of the ensemble: when they all start resembling each other, no additional reduction of the variance of the error is obtained \citep{Ueda1996, Breiman2001}. Our experiments show that using $k=500$ is significantly better than using $k=100$ (p-value is $<$0.001 in a pairwise comparison after Holm's correction) but that the magnitude of the difference is very small. Importantly, however, it shows that, when going from 100 to 500, there is a substantial reduction in the variance in the accuracy between runs. 
In consequence, we make 500 trees the default, as it provides a good trade-off between accuracy and running times.}

\subsection{Contribution of Splitting Functions}
\label{subsec:accuracy_ablative}

\begin{center}
\begin{figure*}[ht]
\centering
\includegraphics[width=0.95\textwidth]{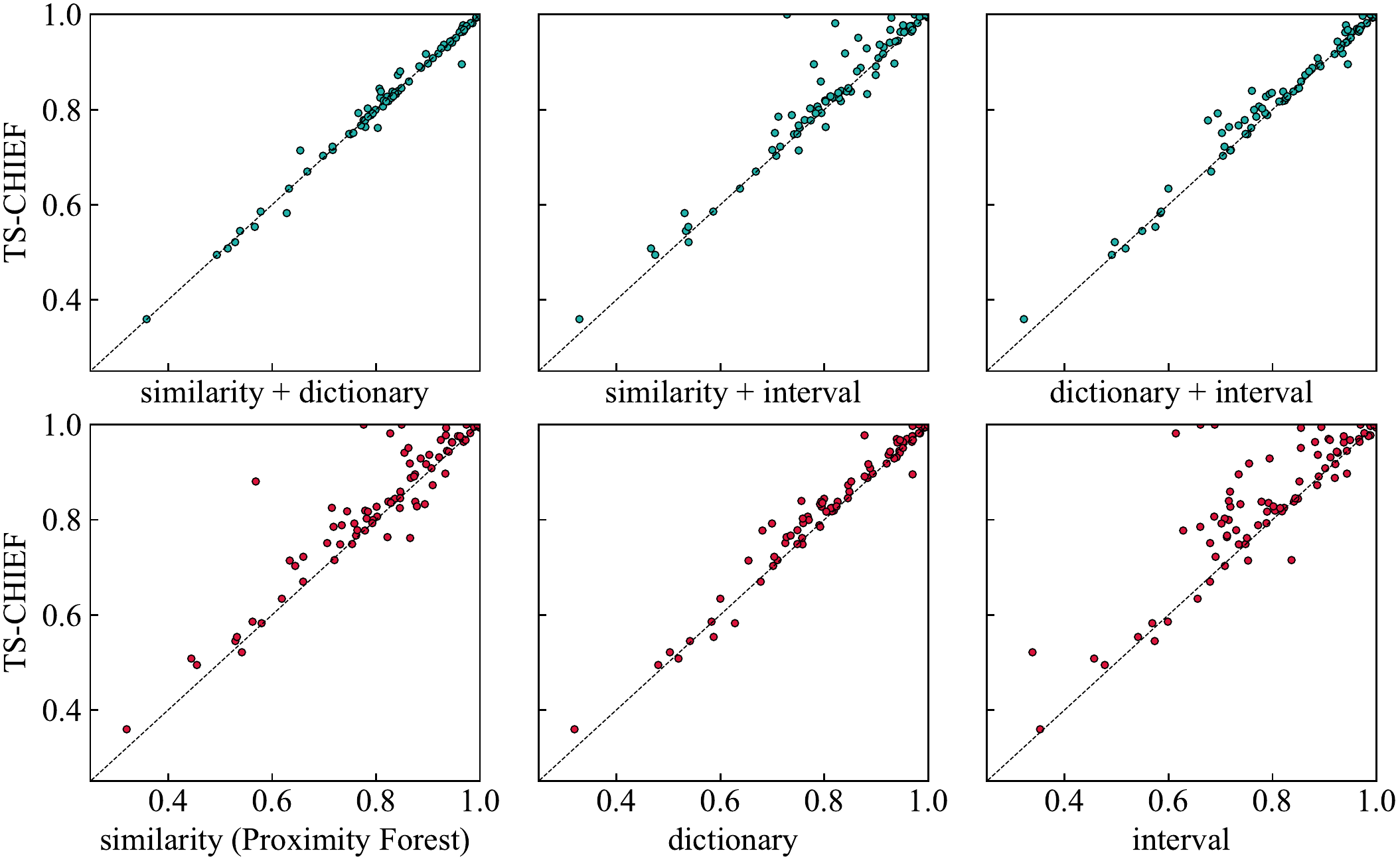}
\vspace*{-5pt}
\caption{Pairwise comparison of accuracy with one \reviewone{(bottom row)} or two \reviewone{(top row)} types of split functions \emph{versus} \ourmethod (where all three types of split functions were used). \reviewone{Similarity \emph{versus} \ourmethod (bottom-left) shows the pairwise comparison of Proximity Forest against \ourmethod.}
}

\label{fig:accu-contributions}
\end{figure*}
\end{center}

\begin{figure}[!ht]
\centering
\includegraphics[width=.95\linewidth]{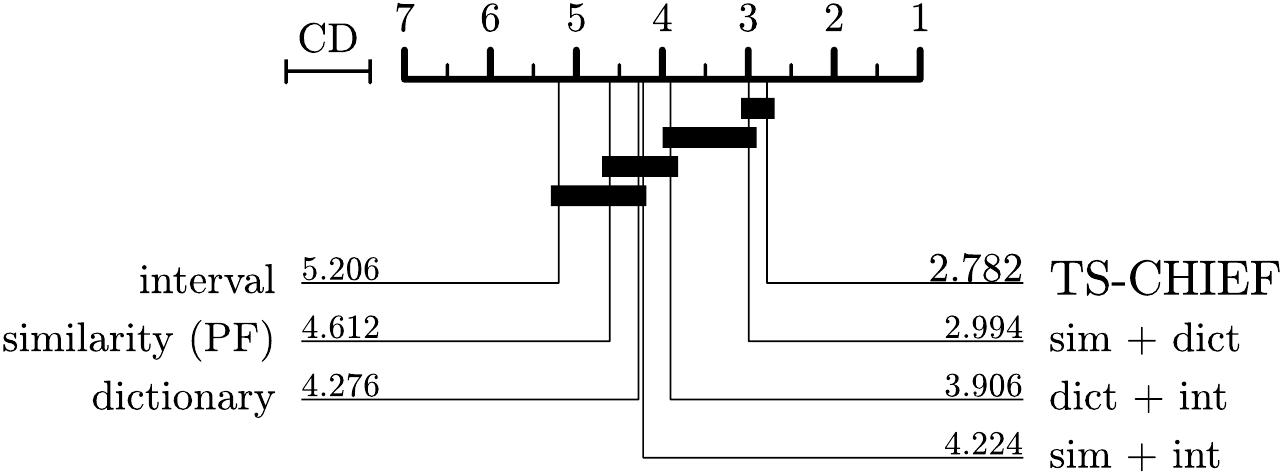}
\vspace*{-5pt}
\caption{
\reviewone{
Critical difference diagram showing the mean ranks of different combinations of split functions.
}}
\label{fig:cd-contributions}
\end{figure}

We also conducted ablation experiments to assess the contribution of each type of splitting function: similarity-based, dictionary-based and interval-based. For this purpose, we assess each variant of \ourmethod created by disabling one of the functions or a pair of the functions.  We performed these experiments with 100 trees, \reviewone{and report the mean accuracy of 10 repetitions.}

Figure~\ref{fig:accu-contributions} displays six scatter-plots comparing the accuracy of \ourmethod using all splitting functions to that of the six ablation configurations. 
The vertical axes indicate the accuracy of \ourmethod with all split functions enabled. The first row compares \ourmethod to variants with a single splitting function disabled \reviewone{(i.e with two types of split functions only)}. The second row compares \ourmethod to variants with only a single splitting function enabled. Please note that the use of only the similarity-based splitting function (first column, second row) corresponds to the \reviewone{Proximity Forest} algorithm \citep{Lucas2018}. Each point indicates one of the 85 UCR datasets. Points above the diagonal dashed line indicate that \ourmethod with all three splitting functions has higher accuracy than the alternative.

The scatter plots on the bottom row indicate that, individually, the  dictionary-based splitter contributes most to the accuracy with 18 wins, 59 losses and 8 ties relative to \ourmethod.  
We can also observe that the magnitudes of its losses tend to be smaller.
Conversely, the interval-based splitter contributes least to the accuracy, with losses of the greatest magnitude relative to \ourmethod. However, it still achieves lower error on 17 datasets, demonstrating that there are some datasets for which the interval-based approach performs well. 

\reviewone{When comparing similarity-based splitter (Proximity Forest) against \ourmethod ($k=100$), the win/draw/loss is 67/2/16 in favor of \ourmethod. There are 5 datasets for which the wins are larger than 10\%: Wine (31\%), ShapeletSim (22\%), OSULeaf (15\%), ECGFiveDays (15\%) and FordB (11\%). When \ourmethod lost, the biggest three losses were for Lighting2 (10\%) Lightning7 (6\%) and FaceAll (5\%).}

In addition, the similarity-based splitter in conjunction with the dictionary-based splitter (that is, the variant with interval-based disabled) is closest to the accuracy of \ourmethod, with 26 wins against \ourmethod, 42 losses and 8 ties.

Figure~\ref{fig:cd-contributions} shows a critical difference diagram summarizing the the relative accuracy of all combinations of the splitting functions.  This confirms our observations from the graphs in Figure \ref{fig:accu-contributions}. The combination of all three types of splitters has the highest average rank. Next come the pairs of splitters, with all pairs outranking the single splitters, albeit marginally for the pair that excludes the dictionary splitter.

\begin{figure}
     \centering
     \begin{subfigure}[b]{.8\textwidth}
         \centering
         \includegraphics[width=\textwidth,height=6.5cm]{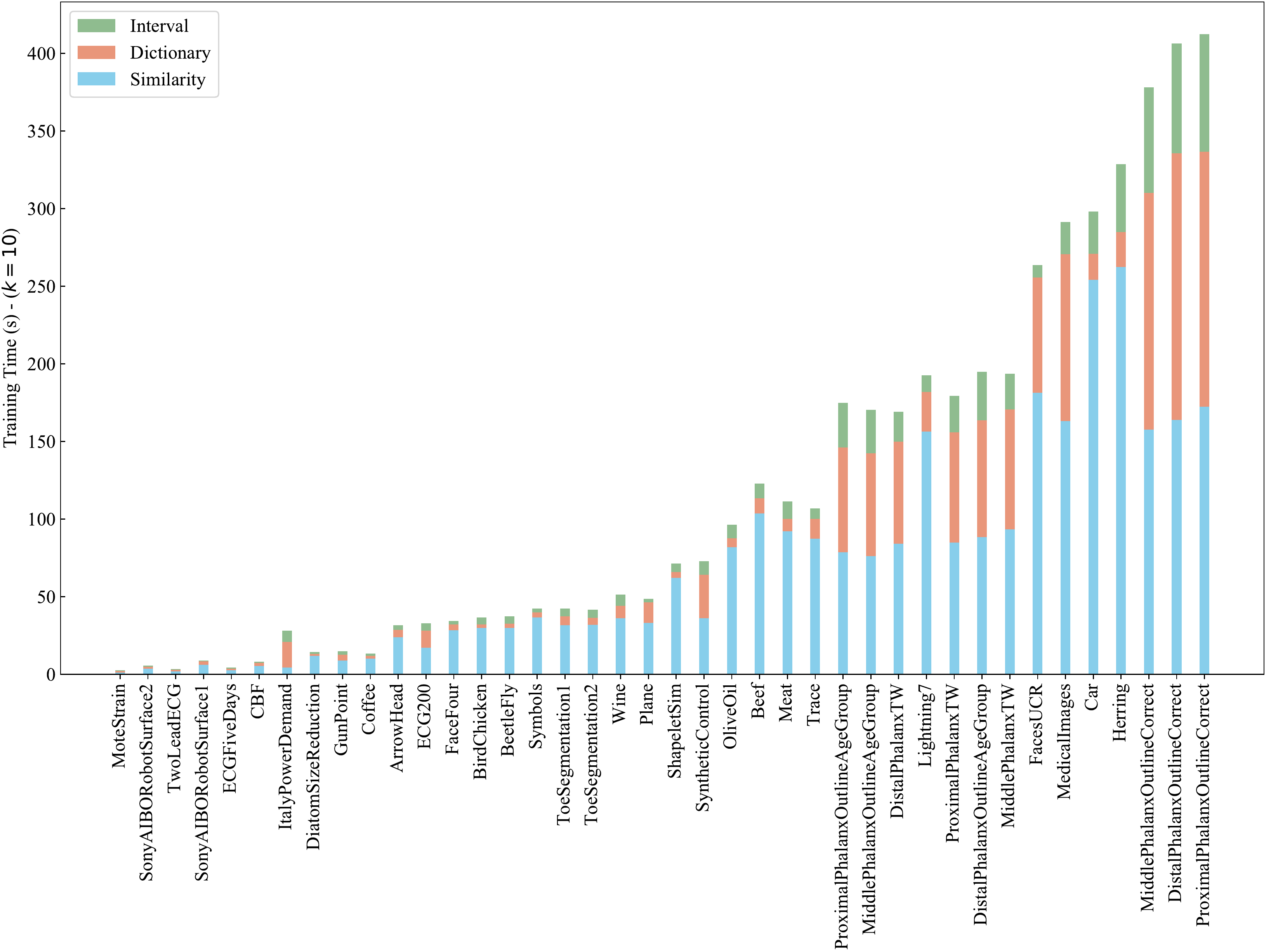}
         \label{fig:splitter-timing-a}
     \end{subfigure}
     \hfill
     \begin{subfigure}[b]{0.8\textwidth}
         \centering
         \includegraphics[width=\textwidth,height=6.5cm]{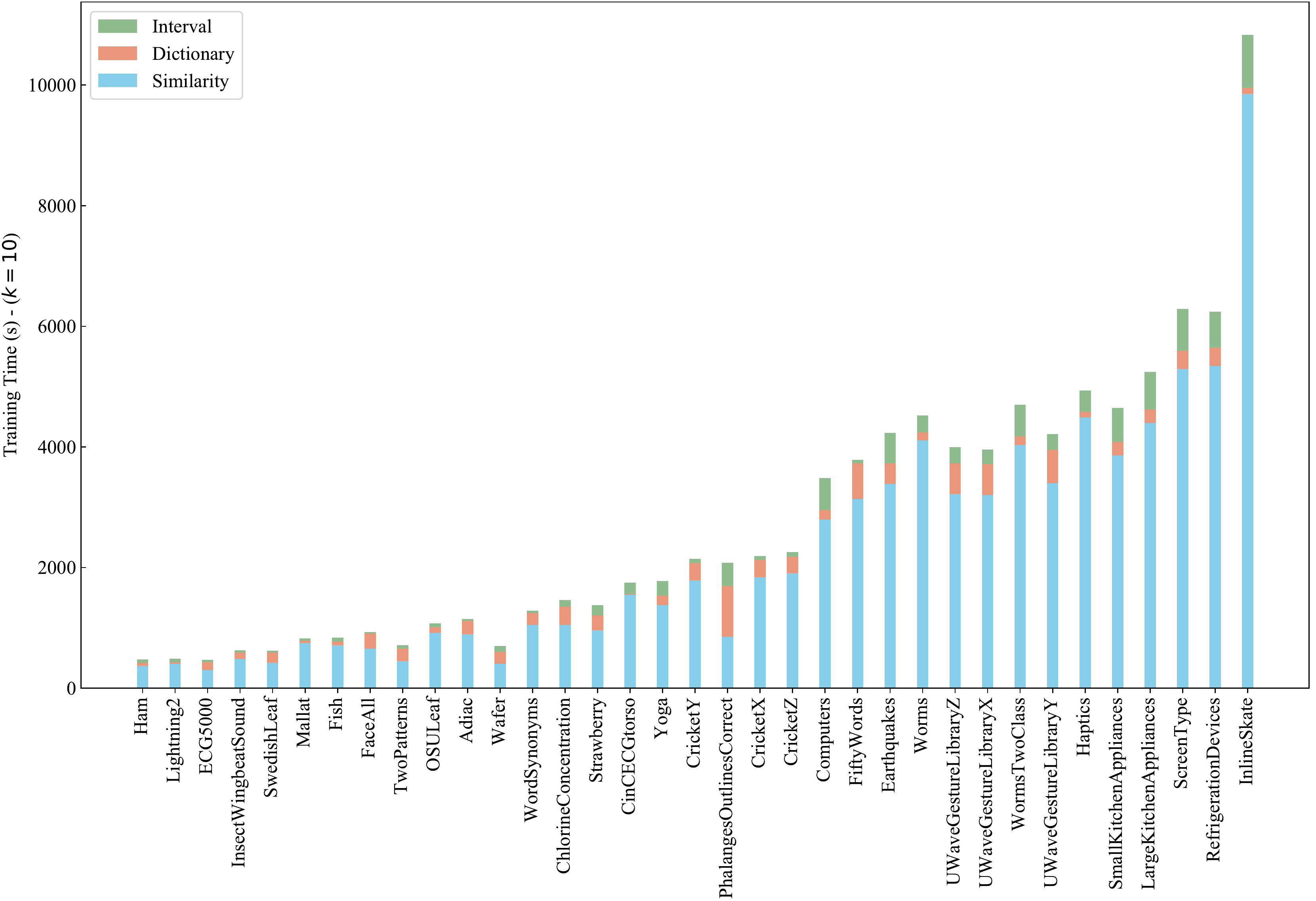}
         \label{fig:splitter-timing-b}
     \end{subfigure}
     \hfill
     \begin{subfigure}[b]{0.8\textwidth}
         \centering
         \includegraphics[width=\textwidth,height=6.5cm]{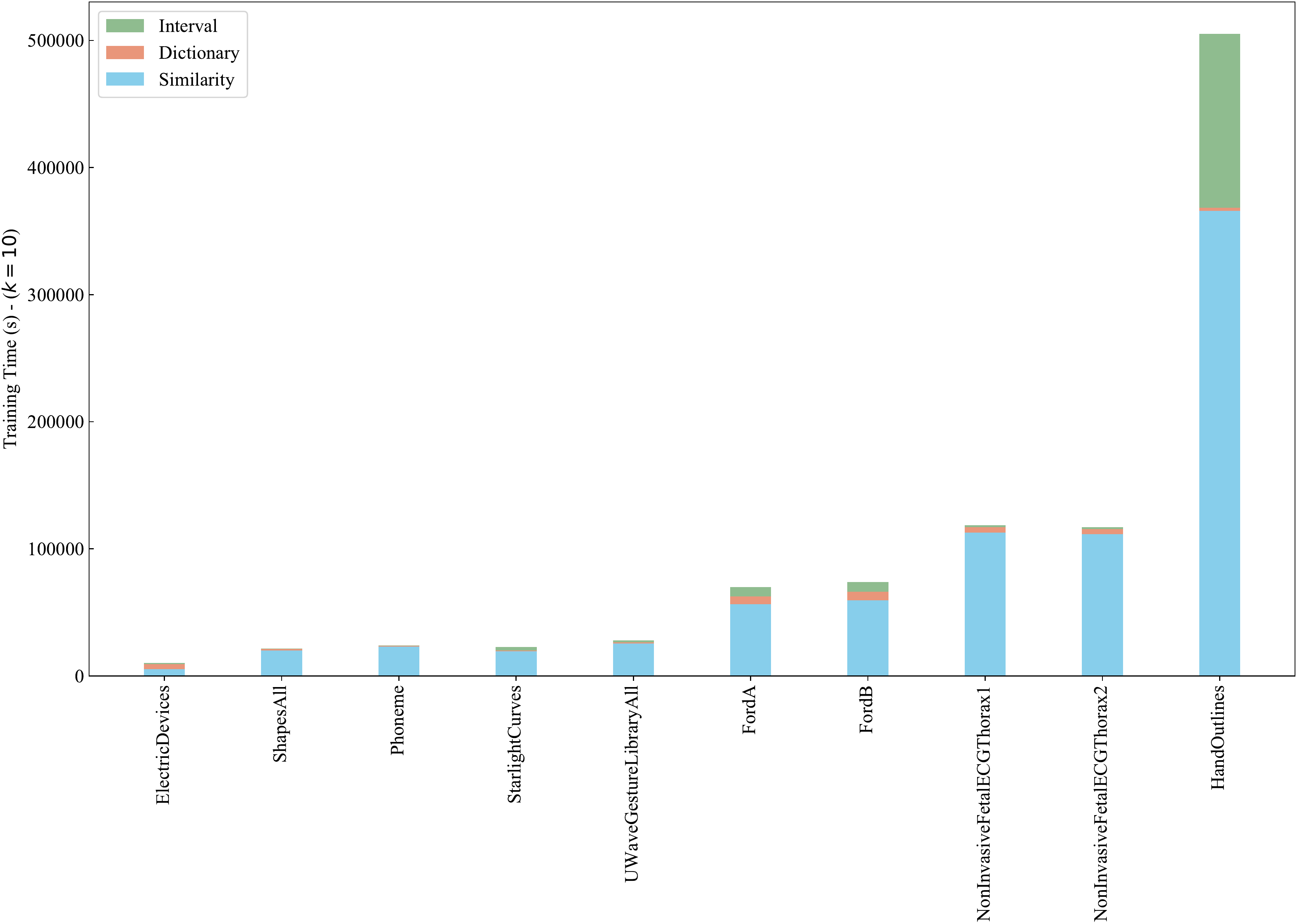}
         \label{fig:splitter-timing-c}
     \end{subfigure}
        \caption{Fraction of training time taken for each splitter type for 85 UCR datasets \citep{UCRArchive2015}. In this experiment, we selected the hyperparameters as follows: number of similarity-based splitters $C_e = 5$, number of dictionary-based splitters $C_b = 100$, and the number of interval-based splitters $C_r=100$. 
We ran this experinment with $k=10$ trees to evaluate the fraction of training time used by each splitter type.}
        \label{fig:splitter-timing}
\end{figure}

\reviewone{
The contribution to accuracy from the interval-based splitter is small, and the sim+dict combination is not statistically different from \ourmethod (p-value is 0.777 in a pairwise comparison after Holm's correction) which uses the three splitters. There are three main reasons why we decided to keep the interval-based splitter in our method. (1) It ranks slightly higher than using only two. (2) It provides a different type of representation which we believe could be useful in real-world applications; in other words, we are conscious that there is a bias in the datasets of the UCR archive and want to prepare our method for unseen datasets as well. (3) Figure~\ref{fig:splitter-timing} (on page~\pageref{fig:splitter-timing})\reviewtwo{, which displays the fraction of time used by each splitting function,} shows that the interval-based splitter takes only a \reviewtwo{small} fraction of the time in \ourmethod, so that the downsides of including it are small.
}

\begin{figure}
     \centering
     \begin{subfigure}[b]{.8\textwidth}
         \centering
         \includegraphics[width=\textwidth,height=6.5cm]{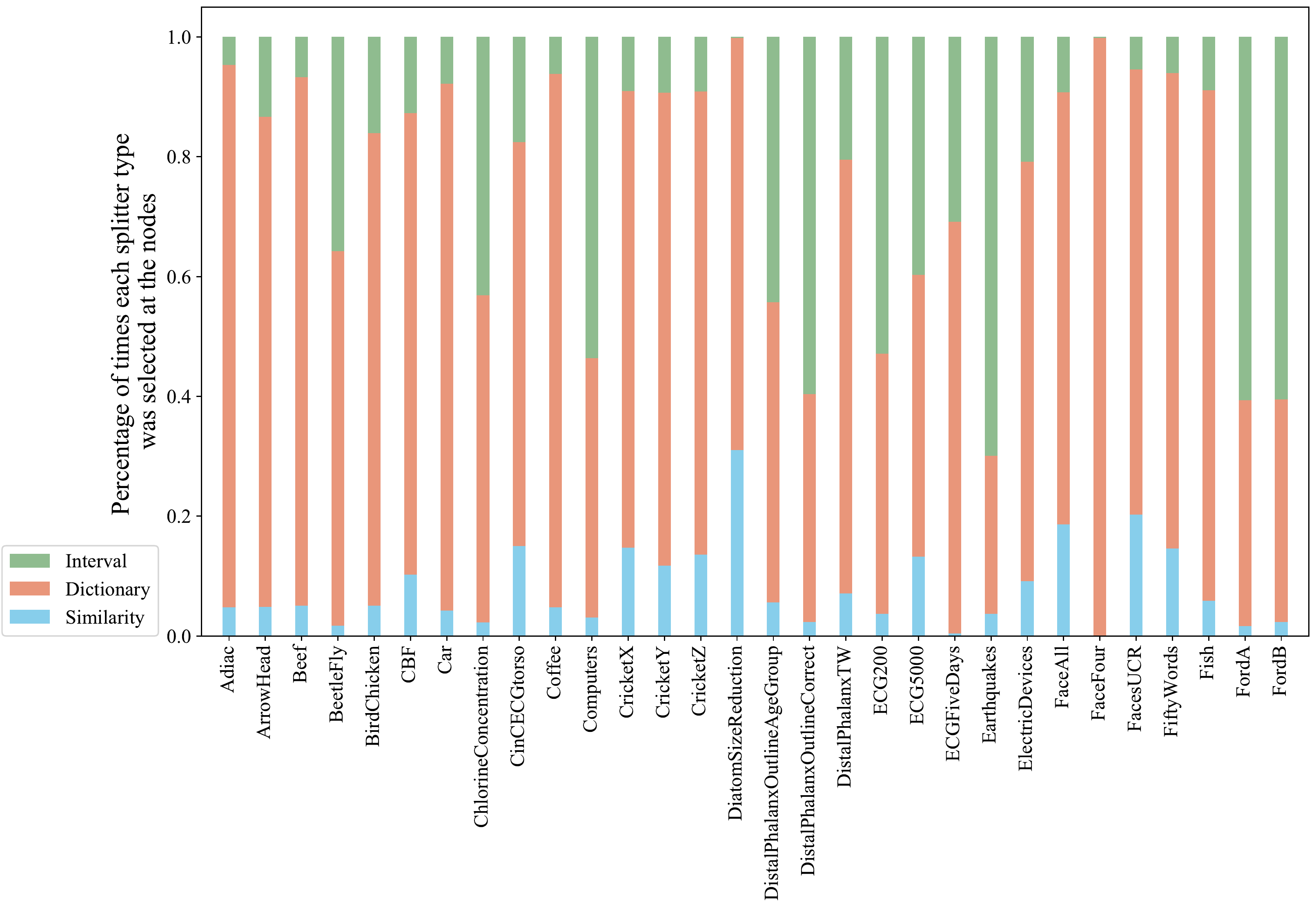}
         \label{fig:splitter-wins-a}
     \end{subfigure}
     \hfill
     \begin{subfigure}[b]{0.8\textwidth}
         \centering
         \includegraphics[width=\textwidth,height=6.5cm]{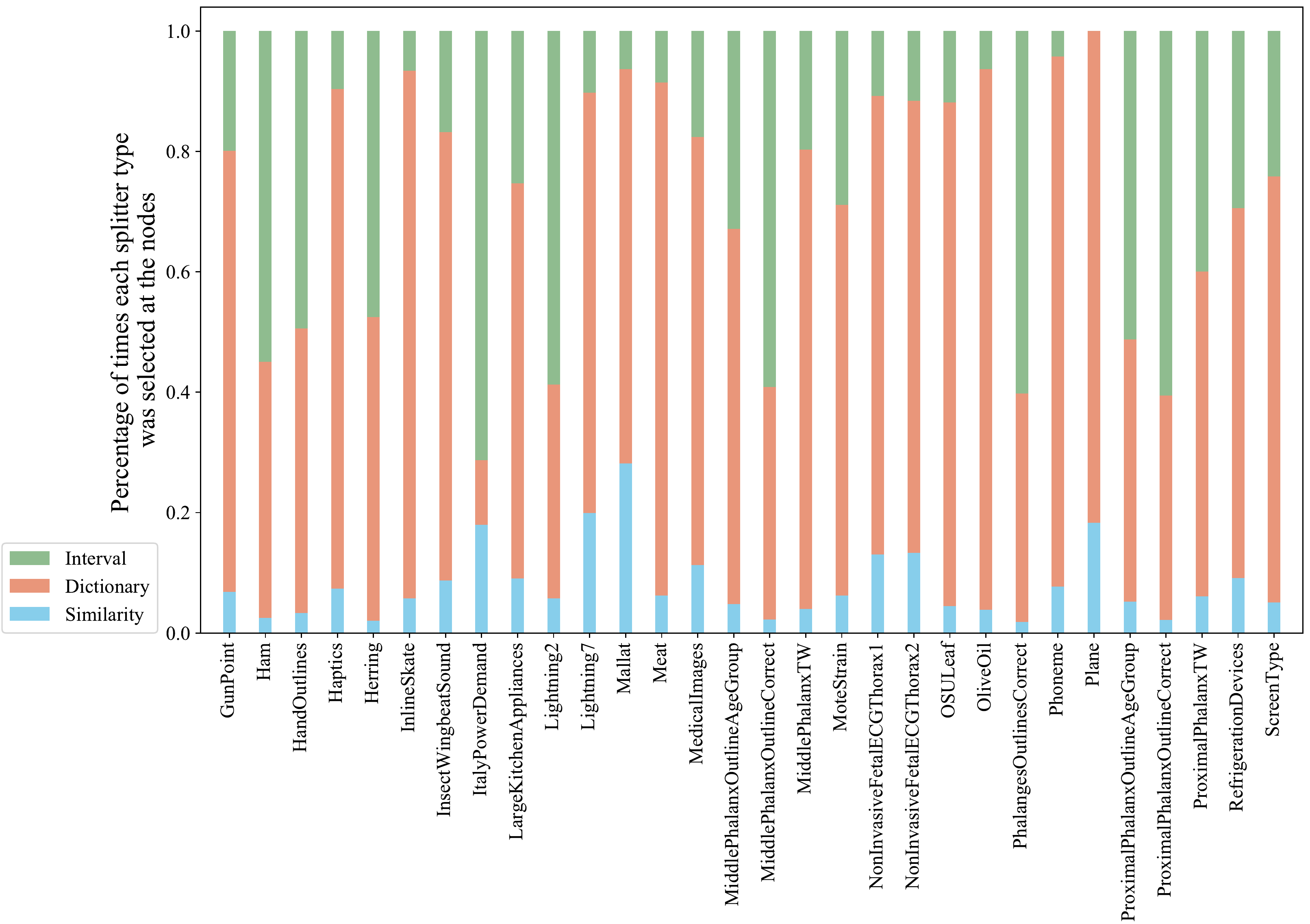}
         \label{fig:splitter-wins-b}
     \end{subfigure}
     \hfill
     \begin{subfigure}[b]{0.8\textwidth}
         \centering
         \includegraphics[width=\textwidth,height=6.5cm]{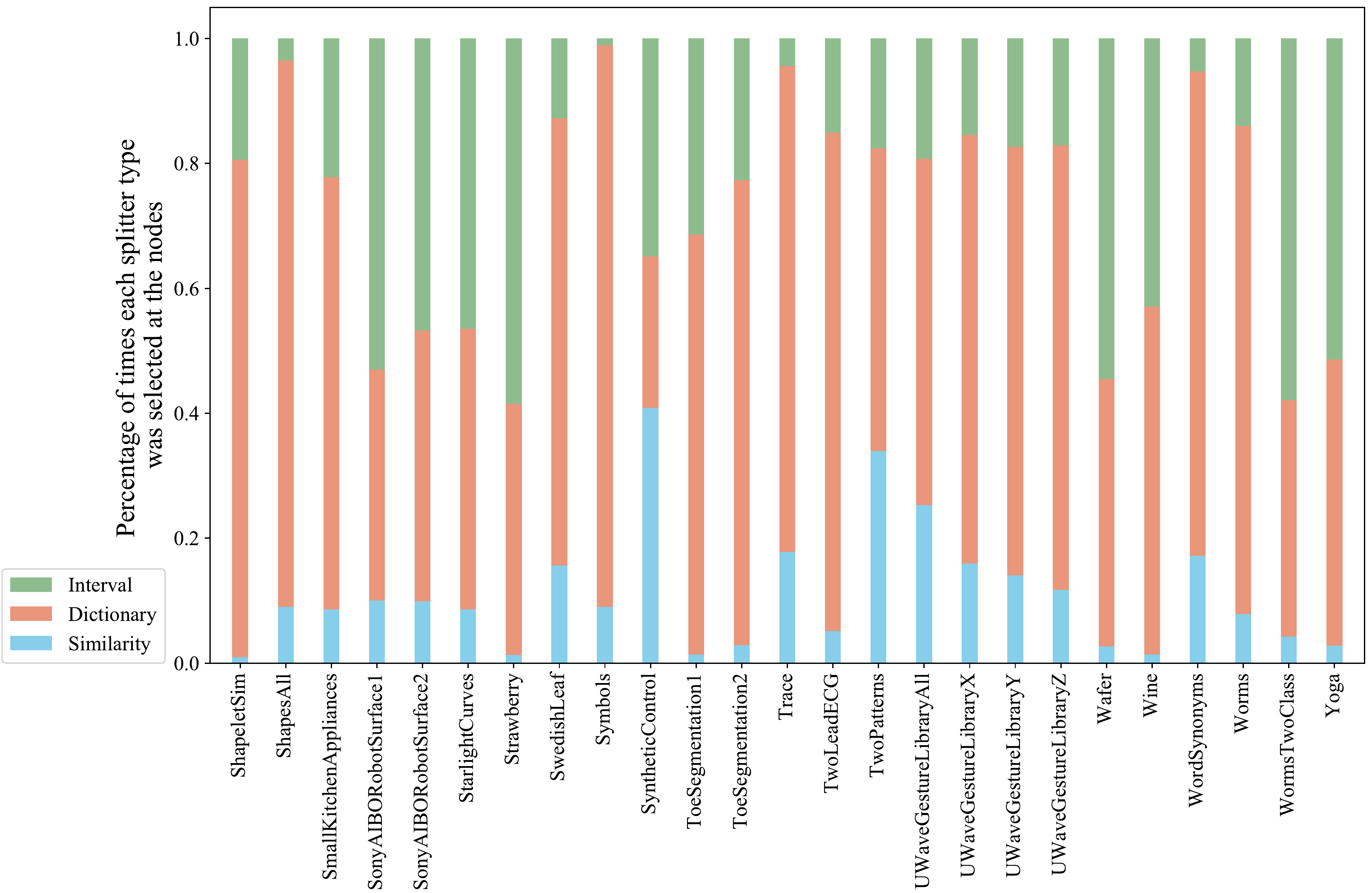}
         \label{fig:splitter-wins-c}
     \end{subfigure}
        \caption{Percentage of times each splitter type was selected at the nodes.
        }
        \label{fig:splitter-wins}
\end{figure}

\reviewone{
To analyze further, Figure~\ref{fig:splitter-wins} (on page~\pageref{fig:splitter-wins}) displays the percentage of times each splitter type was selected at a node. We observe that the dictionary-based splitter ($C_b = 100$) is selected more often than the other two types of splitters, with an average of 60\% of the time, across the 85 datasets. We used $C_e = 5$ for similarity-based splitters, but we also observe that similarity-based splitters were selected 30\% of the time, whereas, an interval-based splitter ($C_r = 100$) was selected only 10\% of the time. It is interesting that, despite that a dictionary-splitter was selected more often, it uses less time (15\%) than the similarity-based splitter (80\%) -- this can be seen from Figure~\ref{fig:splitter-timing}.
}

\subsection{Memory Usage}
\label{subsec:memory}

\reviewone{
    In Section~\ref{subsec:complexity} we saw that the memory complexity of \ourmethod is $O(n \cdot \ell +  k\cdot n\cdot c + t\cdot n \cdot \ell)$. Recall that $t$ is the number of BOSS transformations precomputed at the beginning of training. There is a memory \emph{vs} computational time tradeoff between precomputing $t$ BOSS transformations at the forest level and computing a random BOSS transformation at the tree or node level. To measure the actual memory usage due to the storage of BOSS transformations, we conducted an experiment using $k=1$ on the longest UCR dataset \emph{HandOutlines} ($n$ = 1,000, $\ell$ = 2,700) and on 131k instances (same amount used in Figure~\ref{fig:scale-sat-time}) of SITS dataset (see Section~\ref{subsec:scalability}). We found that the \emph{HandOutlines} uses 36.9~GB and SITS uses 49.8~GB of memory.
    Thus, our decision to precompute BOSS transformations at forest level is due to the following reasons: (1) memory usage is reasonable compared to the computational overhead of transforming at tree or node level, (2) a pool of transformations at the forest level will allow any tree to select any of the $t$ transformations, which helps to improve diversity of the ensemble, whereas, if using, for example, one random BOSS transformation per tree, each tree is restricted to learn from one (or a less diverse pool, if using more than one) transformation.  
}

\section{Conclusions}
\label{sec:conclusion}
We have introduced \ourmethod, which is a scalable and highly accurate algorithm for TSC. We have shown that \ourmethod makes the most of the quasi-linear scalability of trees \reviewone{relative to quantity of data}, together with the last decade of research into deriving accurate representations of time series. 
Our experiments carried out on 85 datasets show that our algorithm reaches state-of-the-art accuracy that rivals  HIVE-COTE, an algorithm which cannot be used in many applications because of its computational complexity. 

We showed that on an application for land-cover mapping, \ourmethod is able to learn a model from 130,000 time series in 2 days, whereas it takes HIVE-COTE 8 days to learn from only 1,500 time series --~a quantity of data from which \ourmethod learns in 13 minutes. \ourmethod offers a general framework for time series classification. We believe that researchers will find it easy to integrate novel transformations and similarity measures and apply them at scale. 

\reviewone{
We conclude by highlighting possible improvements. This includes improving the tradeoff between computation time and memory footprint, incorporating information from different types of potential splitters, as well as finding an automatic way to balance the number of candidate splitters considered for each type (possibly in a manner that is adaptive to the  dataset). Furthermore, future research on \ourmethod could extend it to multivariate time series and datasets with variable-length time series.
}




\section*{Supplementary material}
To ensure reproducibility, a multi-threaded version of this algorithm implemented in Java and the experimental results have been made available in the github repository \url{\gitrepo}.

\section*{Acknowledgements}
This research was supported by the Australian Research Council under grant DE170100037. This material is based upon work supported by the Air Force Office of Scientific Research, Asian Office of Aerospace Research and Development (AOARD) under award number FA2386-17-1-4036.

The authors would like to thank Prof.~Eamonn~Keogh and all the people who have contributed to the UCR time series classification archive.
We also would like to acknowledge the use of source code freely available at \href{http://timeseriesclassification.com/}{http://www.timeseriesclassification.com} and thank Prof.~Anthony~Bagnall and other contributors of the project. We also acknowledge the use of source code freely provided by the original author of BOSS algorithm, Dr.~Patrick~Sch{\"{a}}fer. Finally, we acknowledge the use of two Java libraries \citep{osinski2015hppc, friedmangnu}, which was used to optimize the implementation of our source code.


\begin{thebibliography}{60}
\providecommand{\natexlab}[1]{#1}
\providecommand{\url}[1]{\texttt{#1}}
\expandafter\ifx\csname urlstyle\endcsname\relax
  \providecommand{\doi}[1]{doi: #1}\else
  \providecommand{\doi}{doi: \begingroup \urlstyle{rm}\Url}\fi

\bibitem[Bagnall et~al.(2012)Bagnall, Davis, Hills, and Lines]{Bagnall2012}
A.~Bagnall, L.~Davis, J.~Hills, and J.~Lines.
\newblock {Transformation Based Ensembles for Time Series Classification}.
\newblock \emph{Proceedings of the SIAM Int. Conf. on Data Mining}, pages
  307--318, 2012.

\bibitem[Bagnall et~al.(2015)Bagnall, Lines, Hills, and Bostrom]{Bagnall2016}
A.~Bagnall, J.~Lines, J.~Hills, and A.~Bostrom.
\newblock Time-series classification with {COTE}: the collective of
  transformation-based ensembles.
\newblock \emph{IEEE Transactions on Knowledge and Data Engineering},
  27\penalty0 (9):\penalty0 2522--2535, 2015.

\bibitem[Bagnall et~al.(2017)Bagnall, Lines, Bostrom, Large, and
  Keogh]{Bagnall2017}
A.~Bagnall, J.~Lines, A.~Bostrom, J.~Large, and E.~Keogh.
\newblock {The great time series classification bake off: a review and
  experimental evaluation of recent algorithmic advances}.
\newblock \emph{Data Mining and Knowledge Discovery}, 31\penalty0 (3):\penalty0
  606--660, 2017.

\bibitem[Baydogan and Runger(2016)]{Baydogan2016}
M.~G. Baydogan and G.~Runger.
\newblock {Time series representation and similarity based on local
  autopatterns}.
\newblock \emph{Data Mining and Knowledge Discovery}, 30\penalty0 (2):\penalty0
  476--509, 2016.

\bibitem[Baydogan et~al.(2013)Baydogan, Runger, and Tuv]{Baydogan2013}
M.~G. Baydogan, G.~Runger, and E.~Tuv.
\newblock A bag-of-features framework to classify time series.
\newblock \emph{IEEE transactions on pattern analysis and machine
  intelligence}, 35\penalty0 (11):\penalty0 2796--2802, 2013.

\bibitem[Benavoli et~al.(2016)Benavoli, Corani, and Mangili]{Benavoli2016}
A.~Benavoli, G.~Corani, and F.~Mangili.
\newblock Should we really use post-hoc tests based on mean-ranks?
\newblock \emph{The Journal of Machine Learning Research}, 17\penalty0
  (1):\penalty0 152--161, 2016.

\bibitem[Bostrom and Bagnall(2015)]{Bostrom2015}
A.~Bostrom and A.~Bagnall.
\newblock Binary shapelet transform for multiclass time series classification.
\newblock In \emph{International Conference on Big Data Analytics and Knowledge
  Discovery}, pages 257--269. Springer, 2015.

\bibitem[Breiman(2001)]{Breiman2001}
L.~Breiman.
\newblock {Random forests}.
\newblock \emph{Machine Learning}, 45\penalty0 (1):\penalty0 5--32, 2001.
\newblock ISSN 08856125.

\bibitem[Chen and Ng(2004)]{Chen2004}
L.~Chen and R.~Ng.
\newblock {On The Marriage of Lp-norms and Edit Distance}.
\newblock In \emph{Proceedings of the 13th Int. Conf. on Very Large Data Bases
  (VLDB)}, pages 792--803, 2004.

\bibitem[Chen et~al.(2015)Chen, Keogh, Hu, Begum, Bagnall, Mueen, and
  Batista]{UCRArchive2015}
Y.~Chen, E.~Keogh, B.~Hu, N.~Begum, A.~Bagnall, A.~Mueen, and G.~Batista.
\newblock The {UCR} time series classification archive, July 2015.
\newblock \url{www.cs.ucr.edu/~eamonn/time_series_data/}.

\bibitem[Dau et~al.(2018{\natexlab{a}})Dau, Bagnall, Kamgar, Yeh, Zhu,
  Gharghabi, Ratanamahatana, and Keogh]{Dau2018a}
H.~A. Dau, A.~Bagnall, K.~Kamgar, C.-C.~M. Yeh, Y.~Zhu, S.~Gharghabi, C.~A.
  Ratanamahatana, and E.~Keogh.
\newblock The {UCR} time series archive.
\newblock \emph{arXiv preprint arXiv:1810.07758}, October 2018{\natexlab{a}}.
\newblock \url{https://www.cs.ucr.edu/~eamonn/time_series_data_2018/}.

\bibitem[Dau et~al.(2018{\natexlab{b}})Dau, Keogh, Kamgar, Yeh, Zhu, Gharghabi,
  Ratanamahatana, Yanping, Hu, Begum, Bagnall, Mueen, and
  Batista]{UCRArchive2018}
H.~A. Dau, E.~Keogh, K.~Kamgar, C.-C.~M. Yeh, Y.~Zhu, S.~Gharghabi, C.~A.
  Ratanamahatana, Yanping, B.~Hu, N.~Begum, A.~Bagnall, A.~Mueen, and
  G.~Batista.
\newblock The {UCR} time series classification archive, October
  2018{\natexlab{b}}.
\newblock \url{https://www.cs.ucr.edu/~eamonn/time_series_data_2018/}.

\bibitem[Dem{\v{s}}ar(2006)]{Demsar2006}
J.~Dem{\v{s}}ar.
\newblock {Statistical Comparisons of Classifiers over Multiple Data Sets}.
\newblock \emph{Journal of Machine Learning Research}, 7:\penalty0 1--30, 2006.

\bibitem[Deng et~al.(2013)Deng, Runger, Tuv, and Vladimir]{Deng2013}
H.~Deng, G.~Runger, E.~Tuv, and M.~Vladimir.
\newblock A time series forest for classification and feature extraction.
\newblock \emph{Information Sciences}, 239:\penalty0 142--153, 2013.

\bibitem[Ding et~al.(2008)Ding, Trajcevski, Scheuermann, Wang, and
  Keogh]{Ding2008}
H.~Ding, G.~Trajcevski, P.~Scheuermann, X.~Wang, and E.~J. Keogh.
\newblock {Querying and mining of time series data: experimental comparison of
  representations and distance measures}.
\newblock \emph{Proc. of the VLDB Endowment}, 1\penalty0 (2):\penalty0
  1542--1552, 2008.

\bibitem[Esling and Agon(2012)]{esling2012time}
P.~Esling and C.~Agon.
\newblock Time-series data mining.
\newblock \emph{ACM Computing Surveys (CSUR)}, 45\penalty0 (1):\penalty0 12,
  2012.

\bibitem[Fawaz et~al.(2019)Fawaz, Forestier, Weber, Idoumghar, and
  Muller]{Fawaz2018}
H.~I. Fawaz, G.~Forestier, J.~Weber, L.~Idoumghar, and P.-A. Muller.
\newblock Deep learning for time series classification: a review.
\newblock \emph{Data Mining and Knowledge Discovery}, pages 1--47, Mar 2019.

\bibitem[Friedman and Eden(2013)]{friedmangnu}
E.~Friedman and R.~Eden.
\newblock {GNU} {T}rove: High-performance collections library for {J}ava, 2013.
\newblock \url{https://bitbucket.org/trove4j/trove/src/master/}.

\bibitem[G{\'{o}}recki and {\L}uczak(2013)]{Gorecki2013}
T.~G{\'{o}}recki and M.~{\L}uczak.
\newblock {Using derivatives in time series classification}.
\newblock \emph{Data Mining and Knowledge Discovery}, 26\penalty0 (2):\penalty0
  310--331, 2013.
\newblock ISSN 13845810.

\bibitem[Grabocka et~al.(2014)Grabocka, Schilling, Wistuba, and
  Schmidt-Thieme]{Grabocka2014}
J.~Grabocka, N.~Schilling, M.~Wistuba, and L.~Schmidt-Thieme.
\newblock {Learning time-series shapelets}.
\newblock \emph{Proceedings of the 20th ACM SIGKDD international conference on
  Knowledge discovery and data mining - KDD '14}, pages 392--401, 2014.

\bibitem[Hills et~al.(2014)Hills, Lines, Baranauskas, Mapp, and
  Bagnall]{Hills2014}
J.~Hills, J.~Lines, E.~Baranauskas, J.~Mapp, and A.~Bagnall.
\newblock {Classification of time series by shapelet transformation}.
\newblock \emph{Data Mining and Knowledge Discovery}, 28\penalty0 (4):\penalty0
  851--881, 2014.
\newblock ISSN 13845810.

\bibitem[Hirschberg(1977)]{Hirschberg1977}
D.~S. Hirschberg.
\newblock {Algorithms for the Longest Common Subsequence Problem}.
\newblock \emph{Journal of the ACM}, 24\penalty0 (4):\penalty0 664--675, 1977.

\bibitem[Jeong et~al.(2011)Jeong, Jeong, and Omitaomu]{Jeong2011}
Y.~S. Jeong, M.~K. Jeong, and O.~A. Omitaomu.
\newblock {Weighted dynamic time warping for time series classification}.
\newblock \emph{Pattern Recognition}, 44\penalty0 (9):\penalty0 2231--2240,
  2011.

\bibitem[Karlsson et~al.(2016)Karlsson, Papapetrou, and
  Bostr{\"{o}}m]{Karlsson2016}
I.~Karlsson, P.~Papapetrou, and H.~Bostr{\"{o}}m.
\newblock {Generalized random shapelet forests}.
\newblock \emph{Data Mining and Knowledge Discovery}, 30\penalty0 (5):\penalty0
  1053--1085, 2016.

\bibitem[Keogh and Kasetty(2003)]{Keogh2003}
E.~Keogh and S.~Kasetty.
\newblock On the need for time series data mining benchmarks: a survey and
  empirical demonstration.
\newblock \emph{Data Mining and knowledge discovery}, 7\penalty0 (4):\penalty0
  349--371, 2003.

\bibitem[Keogh et~al.(2001)Keogh, Chakrabarti, Pazzani, and
  Mehrotra]{Keogh2001b}
E.~Keogh, K.~Chakrabarti, M.~Pazzani, and S.~Mehrotra.
\newblock Locally adaptive dimensionality reduction for indexing large time
  series databases.
\newblock \emph{ACM Sigmod Record}, 30\penalty0 (2):\penalty0 151--162, 2001.

\bibitem[Keogh and Pazzani(2001)]{Keogh2001a}
E.~J. Keogh and M.~J. Pazzani.
\newblock {Derivative Dynamic Time Warping}.
\newblock \emph{Proceedings of the 2001 SIAM Int. Conf. on Data Mining}, pages
  1--11, 2001.

\bibitem[Large et~al.(2017)Large, Lines, and Bagnall]{Large2017}
J.~Large, J.~Lines, and A.~Bagnall.
\newblock {The Heterogeneous Ensembles of Standard Classification Algorithms
  (HESCA): the Whole is Greater than the Sum of its Parts}.
\newblock pages 1--31, 2017.
\newblock URL \url{http://arxiv.org/abs/1710.09220}.

\bibitem[Large et~al.(2018)Large, Bagnall, Malinowski, and Tavenard]{Large2018}
J.~Large, A.~Bagnall, S.~Malinowski, and R.~Tavenard.
\newblock {From BOP to BOSS and Beyond: Time Series Classification with
  Dictionary Based Classifiers}.
\newblock pages 1--22, 2018.
\newblock URL \url{http://arxiv.org/abs/1809.06751}.

\bibitem[Le~Guennec et~al.(2016)Le~Guennec, Malinowski, and
  Tavenard]{le2016data}
A.~Le~Guennec, S.~Malinowski, and R.~Tavenard.
\newblock Data augmentation for time series classification using convolutional
  neural networks.
\newblock In \emph{ECML/PKDD Workshop on Advanced Analytics and Learning on
  Temporal Data}, 2016.

\bibitem[Lin et~al.(2007)Lin, Keogh, Wei, and Lonardi]{Lin2007}
J.~Lin, E.~Keogh, L.~Wei, and S.~Lonardi.
\newblock {Experiencing SAX: A novel symbolic representation of time series}.
\newblock \emph{Data Mining and Knowledge Discovery}, 15\penalty0 (2):\penalty0
  107--144, 2007.
\newblock ISSN 13845810.

\bibitem[Lin et~al.(2012)Lin, Khade, and Li]{Lin2012}
J.~Lin, R.~Khade, and Y.~Li.
\newblock {Rotation-invariant similarity in time series using bag-of-patterns
  representation}.
\newblock \emph{Journal of Intelligent Information Systems}, 39\penalty0
  (2):\penalty0 287--315, 2012.

\bibitem[Lines and Bagnall(2015)]{Lines2015}
J.~Lines and A.~Bagnall.
\newblock {Time series classification with ensembles of elastic distance
  measures}.
\newblock \emph{Data Mining and Knowledge Discovery}, 29\penalty0 (3):\penalty0
  565--592, 2015.
\newblock ISSN 13845810.

\bibitem[Lines et~al.(2018)Lines, Taylor, and Bagnall]{Lines2018}
J.~Lines, S.~Taylor, and A.~Bagnall.
\newblock Time series classification with hive-cote: The hierarchical vote
  collective of transformation-based ensembles.
\newblock \emph{ACM Transactions on Knowledge Discovery from Data (TKDD)},
  12\penalty0 (5):\penalty0 52, 2018.

\bibitem[Lucas et~al.(2019)Lucas, Shifaz, Pelletier, O'Neill, Zaidi, Goethals,
  Petitjean, and Webb]{Lucas2018}
B.~Lucas, A.~Shifaz, C.~Pelletier, L.~O'Neill, N.~Zaidi, B.~Goethals,
  F.~Petitjean, and G.~I. Webb.
\newblock Proximity {F}orest: an effective and scalable distance-based
  classifier for time series.
\newblock \emph{Data Mining and Knowledge Discovery}, 33\penalty0 (3):\penalty0
  607--635, May 2019.

\bibitem[Marteau(2009)]{Marteau2009}
P.-F. Marteau.
\newblock {Time Warp Edit Distance with Stiffness Adjustment for Time Series
  Matching}.
\newblock \emph{IEEE Trans. on Pattern Analysis and Machine Intelligence},
  31\penalty0 (2):\penalty0 306--318, 2009.

\bibitem[Middlehurst et~al.(2019)Middlehurst, Vickers, and
  Bagnall]{middlehurst2019scalable}
M.~Middlehurst, W.~Vickers, and A.~Bagnall.
\newblock Scalable dictionary classifiers for time series classification.
\newblock \emph{arXiv preprint arXiv:1907.11815}, 2019.

\bibitem[Mueen et~al.(2011)Mueen, Keogh, and Young]{Mueen2011}
A.~Mueen, E.~Keogh, and N.~Young.
\newblock {Logical-shapelets: An Expressive Primitive for Time Series
  Classification}.
\newblock \emph{Proceedings of the 17th ACM SIGKDD international conference on
  Knowledge discovery and data mining - KDD '11}, page 1154, 2011.

\bibitem[Nwe et~al.(2017)Nwe, Dat, and Ma]{nwe2017convolutional}
T.~L. Nwe, T.~H. Dat, and B.~Ma.
\newblock Convolutional neural network with multi-task learning scheme for
  acoustic scene classification.
\newblock In \emph{2017 Asia-Pacific Signal and Information Processing
  Association Annual Summit and Conference (APSIPA ASC)}, pages 1347--1350.
  IEEE, 2017.

\bibitem[Nweke et~al.(2018)Nweke, Teh, Al-Garadi, and Alo]{nweke2018deep}
H.~F. Nweke, Y.~W. Teh, M.~A. Al-Garadi, and U.~R. Alo.
\newblock Deep learning algorithms for human activity recognition using mobile
  and wearable sensor networks: State of the art and research challenges.
\newblock \emph{Expert Systems with Applications}, 105:\penalty0 233--261,
  2018.

\bibitem[Osinski and Weiss(2015)]{osinski2015hppc}
S.~Osinski and D.~Weiss.
\newblock {HPPC}: High performance primitive collections for {J}ava, 2015.
\newblock \url{ https://labs.carrotsearch.com/hppc.html}.

\bibitem[Pelletier et~al.(2019)Pelletier, Webb, and
  Petitjean]{pelletier2019temporal}
C.~Pelletier, G.~I. Webb, and F.~Petitjean.
\newblock Temporal convolutional neural network for the classification of
  satellite image time series.
\newblock \emph{Remote Sensing}, 11\penalty0 (5):\penalty0 523, 2019.

\bibitem[Rajkomar et~al.(2018)Rajkomar, Oren, Chen, Dai, Hajaj, Hardt, Liu,
  Liu, Marcus, Sun, et~al.]{rajkomar2018scalable}
A.~Rajkomar, E.~Oren, K.~Chen, A.~M. Dai, N.~Hajaj, M.~Hardt, P.~J. Liu,
  X.~Liu, J.~Marcus, M.~Sun, et~al.
\newblock Scalable and accurate deep learning with electronic health records.
\newblock \emph{NPJ Digital Medicine}, 1\penalty0 (1):\penalty0 18, 2018.

\bibitem[Rakthanmanon and Keogh(2013)]{Rakthanmanon2013a}
T.~Rakthanmanon and E.~Keogh.
\newblock {Fast Shapelets: A Scalable Algorithm for Discovering Time Series
  Shapelets}.
\newblock \emph{Proceedings of the 2013 SIAM International Conference on Data
  Mining}, pages 668--676, 2013.

\bibitem[Rakthanmanon et~al.(2013)Rakthanmanon, Campana, Mueen, Batista,
  Westover, Zhu, Zakaria, and Keogh]{Rakthanmanon2013b}
T.~Rakthanmanon, B.~Campana, A.~Mueen, G.~Batista, B.~Westover, Q.~Zhu,
  J.~Zakaria, and E.~Keogh.
\newblock Addressing big data time series: Mining trillions of time series
  subsequences under dynamic time warping.
\newblock \emph{ACM Transactions on Knowledge Discovery from Data (TKDD)},
  7\penalty0 (3):\penalty0 10, 2013.

\bibitem[Sch{\"{a}}fer(2015)]{Schafer2015}
P.~Sch{\"{a}}fer.
\newblock {The BOSS is concerned with time series classification in the
  presence of noise}.
\newblock \emph{Data Mining and Knowledge Discovery}, 29\penalty0 (6):\penalty0
  1505--1530, 2015.

\bibitem[Sch{\"{a}}fer(2016)]{Schafer2016}
P.~Sch{\"{a}}fer.
\newblock {Scalable time series classification}.
\newblock \emph{Data Mining and Knowledge Discovery}, 30\penalty0 (5):\penalty0
  1273--1298, 2016.
\newblock ISSN 1573756X.

\bibitem[Sch{\"{a}}fer and H{\"{o}}gqvist(2012)]{Schafer2012}
P.~Sch{\"{a}}fer and M.~H{\"{o}}gqvist.
\newblock {SFA: a symbolic fourier approximation and index for similarity
  search in high dimensional datasets}.
\newblock \emph{Proceedings of the 15th Int. Conf. on Extending Database
  Technology}, pages 516--527, 2012.

\bibitem[Sch{\"{a}}fer and Leser(2017)]{Schafer2017a}
P.~Sch{\"{a}}fer and U.~Leser.
\newblock {Fast and Accurate Time Series Classification with WEASEL}.
\newblock In \emph{Proceedings of the 2017 ACM on Conf. on Information and
  Knowledge Management (CIKM)}, pages 637--646, 2017.
\newblock ISBN 9781450349185.

\bibitem[Senin and Malinchik(2013)]{Senin2013}
P.~Senin and S.~Malinchik.
\newblock {SAX-VSM: Interpretable time series classification using SAX and
  vector space model}.
\newblock \emph{Proceedings of IEEE Int. Conf. on Data Mining, ICDM}, pages
  1175--1180, 2013.
\newblock ISSN 15504786.

\bibitem[Silva et~al.(2018)Silva, Giusti, Keogh, and
  Batista]{silva2018speeding}
D.~F. Silva, R.~Giusti, E.~Keogh, and G.~E. Batista.
\newblock Speeding up similarity search under dynamic time warping by pruning
  unpromising alignments.
\newblock \emph{Data Mining and Knowledge Discovery}, 32\penalty0 (4):\penalty0
  988--1016, 2018.

\bibitem[Stefan et~al.(2013)Stefan, Athitsos, and Das]{Stefan2013}
A.~Stefan, V.~Athitsos, and G.~Das.
\newblock {The move-split-merge metric for time series}.
\newblock \emph{IEEE Trans. on Knowledge and Data Engineering}, 25\penalty0
  (6):\penalty0 1425--1438, 2013.
\newblock ISSN 10414347.

\bibitem[Susto et~al.(2018)Susto, Cenedese, and Terzi]{susto2018time}
G.~A. Susto, A.~Cenedese, and M.~Terzi.
\newblock Time-series classification methods: Review and applications to power
  systems data.
\newblock In \emph{Big data application in power systems}, pages 179--220.
  Elsevier, 2018.

\bibitem[Tan et~al.(2017)Tan, Webb, and Petitjean]{Tan2017}
C.~W. Tan, G.~I. Webb, and F.~Petitjean.
\newblock Indexing and classifying gigabytes of time series under time warping.
\newblock In \emph{Proceedings of the 2017 SIAM Int. Conf. on Data Mining},
  pages 282--290. SIAM, 2017.

\bibitem[Ueda and Nakano(1996)]{Ueda1996}
N.~Ueda and R.~Nakano.
\newblock Generalization error of ensemble estimators.
\newblock In \emph{IEEE Int. Conf. on Neural Networks}, volume~1, pages 90--95.
  IEEE, 1996.

\bibitem[Wang et~al.(2013)Wang, Liu, She, Nahavandi, and Kouzani]{wang2013bag}
J.~Wang, P.~Liu, M.~F. She, S.~Nahavandi, and A.~Kouzani.
\newblock Bag-of-words representation for biomedical time series
  classification.
\newblock \emph{Biomedical Signal Processing and Control}, 8\penalty0
  (6):\penalty0 634--644, 2013.

\bibitem[Wang et~al.(2019)Wang, Chen, Hao, Peng, and Hu]{wang2018deep}
J.~Wang, Y.~Chen, S.~Hao, X.~Peng, and L.~Hu.
\newblock Deep learning for sensor-based activity recognition: A survey.
\newblock \emph{Pattern Recognition Letters}, 119:\penalty0 3--11, 2019.

\bibitem[Wang et~al.(2017)Wang, Yan, and Oates]{Wang2017}
Z.~Wang, W.~Yan, and T.~Oates.
\newblock Time series classification from scratch with deep neural networks: A
  strong baseline.
\newblock In \emph{2017 International joint conference on neural networks
  (IJCNN)}, pages 1578--1585. IEEE, 2017.

\bibitem[Yang and Wu(2006)]{yang200610}
Q.~Yang and X.~Wu.
\newblock 10 challenging problems in data mining research.
\newblock \emph{International Journal of Information Technology \& Decision
  Making}, 5\penalty0 (04):\penalty0 597--604, 2006.

\bibitem[Ye and Keogh(2009)]{Ye2009}
L.~Ye and E.~Keogh.
\newblock {Time series shapelets}.
\newblock \emph{Proceedings of the 15th ACM SIGKDD Int. Conf. on Knowledge
  Discovery and Data Mining - KDD '09}, page 947, 2009.

\end{thebibliography}

\newpage

\section*{Appendix}


 \begin{table}[!pth]
    \begin{threeparttable}[b]
    \small
    \setlength{\tabcolsep}{3pt}
    \caption{
            \reviewtwo{Complexities of the methods mentioned in Section~\ref{sec:relatedwork}. For tree-based methods, we present the average case complexity.
            \\ 
            Parameters used in this table are: $n$ training size, $\ell$ series length, $c$ no. classes, $w$ window size, $k$ number of trees, $C_e$ no. candidate splits, $e$ max. no. iterations, $\phi$ shapelet scale, $f$ SFA word length, $R$ no. of subseries.}
        }
    \centering
        \begin{tabular}{ p{1.8cm} p{2.9cm} p{2.4cm} p{3.7cm}}
         \hline
         Method & Train Complexity & Test Complexity & Comments \\ [0.5ex] 
         \hline\hline
         \multicolumn{4}{l}{2.1 Similarity-based}\\
         1-NN~DTW (CV) &
         $O(n^2 \cdot \ell^3)$ &
         $O(n \cdot \ell \cdot w)$  &
         \citet{Bagnall2017}, CV: cross-validating all window sizes without using lower bounds \\
         EE & 
         $O(n^2 \cdot \ell^2)$ & 
         $O(n \cdot \ell ^2)$ & 
         \citet{Lines2015} and \citet[Tab.~1]{Bagnall2017} (EE cross-validates 100 parameters) \\ 
         PF & 
         $O(k \cdot n \cdot log(n) \cdot C_e \cdot c \cdot \ell^2)$ & 
         $O(k \cdot log(n) \cdot c \cdot \ell^2) $ & 
         \citet{Lucas2018}  \\ 
         \hline
         \multicolumn{4}{l}{2.2 Interval-based}\\
         RISE & $O(k \cdot n \cdot log(n) \cdot \ell^2)$ & $O(k \cdot log(n) \cdot \ell^2)$\tnote{\#} & \citet{Lines2018} \\  
         TSF & $O(k \cdot n \cdot log(n) \cdot \ell)$ & $O(k \cdot log(n) \cdot \ell^2)$\tnote{\#} & \citet[Tab.~1]{Bagnall2017} \\
         TSBF & $O(k \cdot n \cdot log(n) \cdot \ell \cdot R)$ & \tnote{*} & \citet[Tab.~1]{Bagnall2017}\\  
         LPS & $O(k \cdot n \cdot log(n) \cdot \ell \cdot R)$ & \tnote{*} & \citet[Tab.~1]{Bagnall2017}\\  
        \hline
        \multicolumn{4}{l}{2.3 Shapelet-based}\\
         ST & $O(n^2 \cdot \ell^4)$ & \tnote{*} & \citet{Hills2014}. Uses a combination of 8 general purpose classifiers to classify \\
         LS & $O(n^2 \cdot \ell^2 \cdot e \cdot \phi)$ & \tnote{*} & \citet[Tab.~1]{Bagnall2017}\\
         FS & $O(n \cdot \ell^2)$ & \tnote{*} & \citet{Rakthanmanon2013b} \\
         GRSF & $O(n^2 \cdot \ell^2 \cdot log (n \ell^2))$ & \tnote{*} & \citet{Karlsson2016}, amortized training time complexity\\
        \hline
         \multicolumn{4}{l}{2.4 Dictionary-based}\\
         BOSS & $O(n^2 \cdot \ell^2)$ & $O(n \cdot \ell)$ & \citet[Section~6]{Schafer2015}\\
         BoP & $O(n \cdot \ell (n - w))$ & \tnote{*} & \citet[Tab.~1]{Bagnall2017}\\
         SAX-VSM & $O(n \cdot \ell (n - w))$ & \tnote{*} & \citet[Tab.~1]{Bagnall2017}\\
         BOSS-VS & $O(n \cdot \ell^{\frac{3}{2}})$ & $O(n)$ & \citet[Tab.~1]{Schafer2016}\\
         WEASEL & $O(min(n\ell^2, c^{(2f)} \cdot n))$ & \tnote{*} & \citet{Schafer2017a}, high space complexity \\ 
        \hline
         \multicolumn{4}{l}{2.5 Combinations of Ensembles}\\
         FLAT-COTE & Bounded by ST & Bounded by EE & Bounded by the slowest algorithm\\
         HIVE-COTE & Bounded by ST & Bounded by EE & Bounded by the slowest algorithm\\
        \hline
          \multicolumn{4}{l}{2.6 Deep Learning}\\
         FCN & \tnote{*} & \tnote{*} & \\ 
         ResNet & \tnote{*} & \tnote{*} & \\ 
        \hline
        \hline
        \label{tab:complexities}       
        \end{tabular}
        \begin{tablenotes}
        \footnotesize
         \item[\#] Indicates that the information is not explicitly stated in the associated paper, but we derived the complexity based on our knowledge of the algorithm
         \item[*] Indicates that the information is not explicitly stated in the associated paper
       \end{tablenotes}
  \end{threeparttable}
\end{table}

\small
\begin{center}
\setlength{\tabcolsep}{5pt}
\begin{longtable}{|l|l|l|l|l|l|l|l|l|}
\caption[]{Accuracy of leading TSC classifiers on 85 UCR datasets. The classifiers are 1-Nearest Neighbour with DTW (labelled DTW), BOSS, PF (Proximity Forest), ST (Shapelet Transform), Residual Neural Network (RN), FLAT-COTE (FCT), HIVE-COTE (HCT), and TS-CHIEF (CHIEF). The last two rows show the number of wins \reviewone{(no. of times ranked at 1)} and average ranking of accuracy (Refer to Figure~\ref{fig:cd-us-vs-tsc}). }
\label{tbl:results} \\

\hline 
\multicolumn{1}{|l|}{\textbf{Dataset}} & 
\multicolumn{1}{l|}{\textbf{DTW}} &
\multicolumn{1}{l|}{\textbf{BOSS}}&
\multicolumn{1}{l|}{\textbf{ST}}&
\multicolumn{1}{l|}{\textbf{PF}}& 
\multicolumn{1}{l|}{\textbf{RN}}& 
\multicolumn{1}{l|}{\textbf{FCT}}& 
\multicolumn{1}{l|}{\textbf{HCT}}& 
\multicolumn{1}{l|}{\textbf{CHIEF}} \\ 
\hline 
\endfirsthead

\multicolumn{9}{c}%
{{\bfseries \tablename\ \thetable{} -- continued from previous page}} \\
\hline 
\multicolumn{1}{|l|}{\textbf{Dataset}} & 
\multicolumn{1}{l|}{\textbf{DTW}} &
\multicolumn{1}{l|}{\textbf{BOSS}}&
\multicolumn{1}{l|}{\textbf{ST}}&
\multicolumn{1}{l|}{\textbf{PF}}& 
\multicolumn{1}{l|}{\textbf{RN}}& 
\multicolumn{1}{l|}{\textbf{FCT}}& 
\multicolumn{1}{l|}{\textbf{HCT}}& 
\multicolumn{1}{l|}{\textbf{CHIEF}} \\ 
\hline 
\endhead

\hline \multicolumn{9}{|r|}{{Continued on next page}} \\ \hline
\endfoot

\hline \hline
\endlastfoot

    Adiac & 60.87 & 76.47 & 78.26 & 73.40 & \textbf{82.89} & 79.03 & 81.07 & 79.80 \\
    ArrHead & 80.00 & 83.43 & 73.71 & \textbf{87.54} & 84.46 & 81.14 & 86.29 & 83.27 \\
    Beef  & 66.67 & 80.00 & 90.00 & 72.00 & 75.33 & 86.67 & \textbf{93.33} & 70.61 \\
    BeetleFly & 65.00 & 90.00 & 90.00 & 87.50 & 85.00 & 80.00 & \textbf{95.00} & 91.36 \\
    BirdChi & 70.00 & \textbf{95.00} & 80.00 & 86.50 & 88.50 & 90.00 & 85.00 & 90.91 \\
    CBF   & 99.44 & 99.78 & 97.44 & 99.33 & 99.50 & 99.56 & \textbf{99.89} & 99.79 \\
    Car   & 76.67 & 83.33 & 91.67 & 84.67 & \textbf{92.50} & 90.00 & 86.67 & 85.45 \\
    ChConc & 65.00 & 66.09 & 69.97 & 63.39 & \textbf{84.36} & 72.71 & 71.20 & 71.67 \\
    CinCECGT & 93.04 & 88.70 & 95.43 & 93.43 & 82.61 & 99.49 & \textbf{99.64} & 98.32 \\
    Coffee & \textbf{100.0} & \textbf{100.0} & 96.43 & \textbf{100.0} & \textbf{100.0} & \textbf{100.0} & \textbf{100.0} & \textbf{100.0} \\
    Comp  & 62.40 & 75.60 & 73.60 & 64.44 & \textbf{81.48} & 74.00 & 76.00 & 70.51 \\
    CricketX & 77.95 & 73.59 & 77.18 & 80.21 & 79.13 & 80.77 & \textbf{82.31} & 81.38 \\
    CricketY & 75.64 & 75.38 & 77.95 & 79.38 & 80.33 & 82.56 & \textbf{84.87} & 80.19 \\
    CricketZ & 73.59 & 74.62 & 78.72 & 80.10 & 81.15 & 81.54 & 83.08 & \textbf{83.40} \\
    DiaSzRed & 93.46 & 93.14 & 92.48 & 96.57 & 30.13 & 92.81 & 94.12 & \textbf{97.30} \\
    DiPhOAG & 62.59 & 74.82 & \textbf{76.98} & 73.09 & 71.65 & 74.82 & 76.26 & 74.62 \\
    DiPhOC & 72.46 & 72.83 & 77.54 & \textbf{79.28} & 77.10 & 76.09 & 77.17 & 78.23 \\
    DiPhTW & 63.31 & 67.63 & 66.19 & 65.97 & 66.47 & \textbf{69.78} & 68.35 & 67.04 \\
    ECG200 & 88.00 & 87.00 & 83.00 & \textbf{90.90} & 87.40 & 88.00 & 85.00 & 86.18 \\
    ECG5000 & 92.51 & 94.13 & 94.38 & 93.65 & 93.42 & 94.60 & \textbf{94.62} & 94.54 \\
    ECG5D & 79.67 & \textbf{100.0} & 98.37 & 84.92 & 97.48 & 99.88 & \textbf{100.0} & \textbf{100.0} \\
    Earthqua & 72.66 & 74.82 & 74.10 & \textbf{75.40} & 71.15 & 74.82 & 74.82 & 74.82 \\
    ElectDev & 63.08 & \textbf{79.92} & 74.70 & 70.60 & 72.91 & 71.33 & 77.03 & 75.53 \\
    FaceAll & 80.77 & 78.17 & 77.87 & 89.38 & 83.88 & \textbf{91.78} & 80.30 & 84.14 \\
    FaceFour & 89.77 & \textbf{100.0} & 85.23 & 97.39 & 95.45 & 89.77 & 95.45 & \textbf{100.0} \\
    FacesUCR & 90.78 & 95.71 & 90.59 & 94.59 & 95.47 & 94.24 & 96.29 & \textbf{96.63} \\
    50Words & 76.48 & 70.55 & 70.55 & 83.14 & 73.96 & 79.78 & 80.88 & \textbf{84.50} \\
    Fish  & 83.43 & 98.86 & 98.86 & 93.49 & 97.94 & 98.29 & 98.86 & \textbf{99.43} \\
    FordA & 66.52 & 92.95 & \textbf{97.12} & 85.46 & 92.05 & 95.68 & 96.44 & 94.10 \\
    FordB & 59.88 & 71.11 & 80.74 & 71.49 & \textbf{91.31} & 80.37 & 82.35 & 82.96 \\
    GunPoint & 91.33 & \textbf{100.0} & \textbf{100.0} & 99.73 & 99.07 & \textbf{100.0} & \textbf{100.0} & \textbf{100.0} \\
    Ham   & 60.00 & 66.67 & 68.57 & 66.00 & \textbf{75.71} & 64.76 & 66.67 & 71.52 \\
    HandOut & 87.84 & 90.27 & \textbf{93.24} & 92.14 & 91.11 & 91.89 & \textbf{93.24} & 93.22 \\
    Haptics & 41.56 & 46.10 & \textbf{52.27} & 44.45 & 51.88 & \textbf{52.27} & 51.95 & 51.68 \\
    Herring & 53.12 & 54.69 & 67.19 & 57.97 & 61.88 & 62.50 & \textbf{68.75} & 58.81 \\
    InlSkate & 38.73 & 51.64 & 37.27 & \textbf{54.18} & 37.31 & 49.45 & 50.00 & 52.69 \\
    InWSnd & 57.37 & 52.32 & 62.68 & 61.87 & 50.65 & 65.25 & \textbf{65.51} & 64.29 \\
    ItPwDem & 95.53 & 90.86 & 94.75 & 96.71 & 96.30 & 96.11 & 96.31 & \textbf{97.06} \\
    LKitApp & 79.47 & 76.53 & 85.87 & 78.19 & \textbf{89.97} & 84.53 & 86.40 & 80.68 \\
    Light2 & \textbf{86.89} & 83.61 & 73.77 & 86.56 & 77.05 & \textbf{86.89} & 81.97 & 74.81 \\
    Light7 & 71.23 & 68.49 & 72.60 & 82.19 & \textbf{84.52} & 80.82 & 73.97 & 76.34 \\
    Mallat & 91.43 & 93.82 & 96.42 & 95.76 & 97.16 & 95.39 & 96.20 & \textbf{97.50} \\
    Meat  & 93.33 & 90.00 & 85.00 & 93.33 & \textbf{96.83} & 91.67 & 93.33 & 88.79 \\
    MdImg & 74.74 & 71.84 & 66.97 & 75.82 & 77.03 & 75.79 & 77.76 & \textbf{79.58} \\
    MdPhOAG & 51.95 & 54.55 & \textbf{64.29} & 56.23 & 56.88 & 63.64 & 59.74 & 58.32 \\
    MdPhOC & 76.63 & 78.01 & 79.38 & 83.64 & 80.89 & 80.41 & 83.16 & \textbf{85.35} \\
    MdPhTW & 50.65 & 54.55 & 51.95 & 52.92 & 48.44 & \textbf{57.14} & \textbf{57.14} & 55.02 \\
    MtStrain & 86.58 & 87.86 & 89.70 & 90.24 & 92.76 & 93.69 & 93.29 & \textbf{94.75} \\
    NoECGT1 & 82.90 & 83.82 & \textbf{94.96} & 90.66 & 94.54 & 93.13 & 93.03 & 91.13 \\
    NoECGT2 & 87.02 & 90.08 & \textbf{95.11} & 93.99 & 94.61 & 94.55 & 94.45 & 94.50 \\
    OSULeaf & 59.92 & 95.45 & 96.69 & 82.73 & 97.85 & 96.69 & 97.93 & \textbf{99.14} \\
    OliveOil & 86.67 & 86.67 & \textbf{90.00} & 86.67 & 83.00 & \textbf{90.00} & \textbf{90.00} & 88.79 \\
    PhalanOC & 76.11 & 77.16 & 76.34 & 82.35 & 83.90 & 77.04 & 80.65 & \textbf{84.50} \\
    Phoneme & 22.68 & 26.48 & 32.07 & 32.01 & 33.43 & 34.92 & \textbf{38.24} & 36.91 \\
    Plane & \textbf{100.0} & \textbf{100.0} & \textbf{100.0} & \textbf{100.0} & \textbf{100.0} & \textbf{100.0} & \textbf{100.0} & \textbf{100.0} \\
    PrxPhOAG & 78.54 & 83.41 & 84.39 & 84.63 & 85.32 & 85.37 & \textbf{85.85} & 84.97 \\
    PrxPhOC & 79.04 & 84.88 & 88.32 & 87.32 & \textbf{92.13} & 86.94 & 87.97 & 88.82 \\
    PrxPhTW & 76.10 & 80.00 & 80.49 & 77.90 & 78.05 & 78.05 & 81.46 & \textbf{81.86} \\
    RefDev & 44.00 & 49.87 & \textbf{58.13} & 53.23 & 52.53 & 54.67 & 55.73 & 55.83 \\
    ScrType & 41.07 & 46.40 & 52.00 & 45.52 & \textbf{62.16} & 54.67 & 58.93 & 50.81 \\
    ShpSim & 69.44 & \textbf{100.0} & 95.56 & 77.61 & 77.94 & 96.11 & \textbf{100.0} & \textbf{100.0} \\
    ShpAll & 80.17 & 90.83 & 84.17 & 88.58 & 92.13 & 89.17 & 90.50 & \textbf{93.00} \\
    SKitApp & 67.20 & 72.53 & 79.20 & 74.43 & 78.61 & 77.60 & \textbf{85.33} & 82.21 \\
    SonyRS1 & 69.55 & 63.23 & 84.36 & 84.58 & \textbf{95.81} & 84.53 & 76.54 & 82.64 \\
    SonyRS2 & 85.94 & 85.94 & 93.39 & 89.63 & \textbf{97.78} & 95.17 & 92.76 & 92.48 \\
    StarCurv & 89.83 & 97.78 & 97.85 & 98.13 & 97.18 & 97.96 & 98.15 & \textbf{98.24} \\
    Strwbe & 94.59 & 97.57 & 96.22 & 96.84 & \textbf{98.05} & 95.14 & 97.03 & 96.63 \\
    SwdLeaf & 84.64 & 92.16 & 92.80 & 94.66 & 95.63 & 95.52 & 95.36 & \textbf{96.55} \\
    Symbols & 93.77 & 96.68 & 88.24 & 96.16 & 90.64 & 96.38 & 97.39 & \textbf{97.66} \\
    SynCtl & 98.33 & 96.67 & 98.33 & 99.53 & 99.83 & \textbf{100.0} & 99.67 & 99.79 \\
    ToeSeg1 & 75.00 & 93.86 & 96.49 & 92.46 & 96.27 & 97.37 & \textbf{98.25} & 96.53 \\
    ToeSeg2 & 90.77 & \textbf{96.15} & 90.77 & 86.23 & 90.62 & 91.54 & 95.38 & 95.38 \\
    Trace & 99.00 & \textbf{100.0} & \textbf{100.00} & \textbf{100.0} & \textbf{100.00} & \textbf{100.0} & \textbf{100.0} & \textbf{100.0} \\
    2LeadECG & 86.83 & 98.07 & 99.74 & 98.86 & \textbf{100.0} & 99.30 & 99.65 & 99.46 \\
    2Pttrns & 99.85 & 99.30 & 95.50 & 99.96 & 99.99 & \textbf{100.0} & \textbf{100.0} & \textbf{100.0} \\
    UWaAll & 96.23 & 93.89 & 94.22 & \textbf{97.23} & 85.95 & 96.43 & 96.85 & 96.89 \\
    UWaX  & 77.44 & 76.21 & 80.29 & 82.86 & 78.05 & 82.19 & 83.98 & \textbf{84.11} \\
    UWaY  & 70.18 & 68.51 & 73.03 & 76.15 & 67.01 & 75.85 & 76.55 & \textbf{77.23} \\
    UWaZ  & 67.50 & 69.49 & 74.85 & 76.40 & 75.01 & 75.04 & 78.31 & \textbf{78.44} \\
    Wafer & 99.59 & 99.48 & \textbf{100.0} & 99.55 & 99.86 & 99.98 & 99.94 & 99.91 \\
    Wine  & 61.11 & 74.07 & 79.63 & 56.85 & 74.44 & 64.81 & 77.78 & \textbf{89.06} \\
    WordSyn & 74.92 & 63.79 & 57.05 & 77.87 & 62.24 & 75.71 & 73.82 & \textbf{78.74} \\
    Worms & 53.25 & 55.84 & 74.03 & 71.82 & 79.09 & 62.34 & 55.84 & \textbf{80.17} \\
    Worms2C & 58.44 & \textbf{83.12} & \textbf{83.12} & 78.44 & 74.68 & 80.52 & 77.92 & 81.58 \\
    Yoga  & 84.30 & \textbf{91.83} & 81.77 & 87.86 & 87.02 & 87.67 & 91.77 & 83.47 \\
   \hline \hline 
    Avg.Rank & 6.982  & 5.400  & 4.806  & 4.818  & 4.300  & 3.818  & \textbf{2.941}  & 2.935 \\
    No. of times &&&&&&&&\\
    ranked 1  & 3     & 12    & 14     & 9    & 18    & 12    & 23    & \textbf{31} \\
\end{longtable}
\end{center}



\end{document}